\documentclass[3p,times]{elsarticle}

\usepackage{ecrc}

\volume{00}

\firstpage{1}

\journalname{Procedia Computer Science}

\runauth{}

\jid{procs}

\jnltitlelogo{Procedia Computer Science}

\CopyrightLine{2011}{Published by Elsevier Ltd.}

\usepackage{amssymb}

\usepackage[figuresright]{rotating}

\usepackage{moreverb,url}
\usepackage{graphicx}
\usepackage{graphicx}
\usepackage{amsmath}
\usepackage{algorithmic}
\usepackage{algorithm}
\usepackage{graphicx}
\usepackage{multirow}
\usepackage{diagbox}
\usepackage{wrapfig}
\usepackage{amsthm}
\usepackage{subfig}
\usepackage{tikz}
\usepackage{scalefnt}
\usepackage{color}
\usetikzlibrary{shapes.geometric, arrows}
\usetikzlibrary{positioning, shapes.geometric}
\theoremstyle{definition}
\newtheorem{defn}{Definition}
\begin{document}

\begin{frontmatter}



\dochead{}

\title{Intention Recognition for Multiple Agents}


\author[1]{Zhang Zhang}
\author[2]{Yifeng Zeng\corref{cor1}\fnref{fn2}}
\ead{Yifeng.Zeng@northumbria.ac.uk}
\author[3]{Yingke Chen\corref{cor1}}
\ead{Y.Chen@tees.ac.uk}

\cortext[cor1]{Corresponding author}
\fntext[fn2]{The author has equal contribuation with the first author.}
\address[1]{National Engineering Laboratory for Electric Vehicles, Beijing Institute of Technology, Beijing 100081, China}
\address[2]{University of Northumbria, Sutherland Building
Newcastle-upon-Tyne, NE1 8ST, United Kingdom}
\address[3]{University of Teesside, Campus Heart, Southfield Rd, Middlesbrough TS1 3BX, United Kingdom}

\begin{abstract}
Intention recognition is an important step to facilitate collaboration in multi-agent systems. Existing work mainly focuses on intention recognition in a single-agent setting and uses a descriptive model, e.g. Bayesian networks, in the recognition process. In this paper, we resort to a prescriptive approach to model agents' behaviour where which their intentions are hidden in implementing their plans. We introduce  landmarks into the behavioural model therefore enhancing informative features for identifying common intentions for multiple agents.  We further refine the model by focusing only action sequences in their plan and provide a light model for identifying and comparing their intentions. The new models provide a simple approach of grouping agents' common intentions upon partial plans observed in agents' interactions. We provide experimental results in support.
\end{abstract}

\begin{keyword}
Intelligent Agents, Intention Recognition, Decision Making
\end{keyword}

\end{frontmatter}


\section{Introduction}
Intention is one of the most important properties for designing an intelligent agent, and represents the courses of actions that the agent has committed to carry out~\cite{wooldridge2009introduction}. It is implemented as a stack of plan instances when the agent is committed to a goal\cite{DBLP:journals}. Intention recognition is the process of recognising the agent's intention through observing its behaviours and their impact on an environment. The recognition is vital for designing intelligent agents so as to adapt their behaviours in achieving the goals.

The previous research focuses on intention recognition for a single agent and has shown interesting applications in  home-care service for the elderly~\cite{pereira2013context, han2011context}. Little research has been found in addressing the problem of recognising intention for multiple agents. However, there is seen increasing amount of research and applications on multi-agent systems where a task is rather complicated and demands collaboration from multiple agents~\cite{stone2020broader}. Intention recognition for multiple agents would facilitate the system adaptation and improve their collaboration to achieve a common goal. Once the agents are found to share a common intention, their behaviour becomes predictable and they could be guided in a better way to find potential collaboration with other agents.

In a more broad perspective, multi-agent intention recognition would improve the understanding of joint behaviours of multiple agents and advise proper intervention in their plans e.g. designing a timely counter-strategy to avoid crowd attacks~\cite{perrault2019ai}. In this paper, we aim to develop a proper procedure for identifying common intentions for multiple agents.

The common intention discovery can exploit potential collaboration for the agents so as to better achieve their goals. From the perspective of individual agents, a subject agent can infer intentions of other agents based on what it observes from the other agents. This particularly occurs to  the case when the agents can not disclose their intentions to each other or they have limited communication while they are executing their plans to complete a task. On the other hand, from the perspective of an observer to a multi-agent system, he/she  can identify intentions of individual agents from his/her observations about their behaviours in the system. We consider a fully observable environment~(Markov decision process~(MDP)~\cite{bellman1957markovian} like case) where environmental states and agents' actions can be fully observed~(with perfect information) by others in a problem domain. To identify their intentions, we need to discover how the observed behaviours are reflected in their full plans that they are following in order to complete their tasks. Unfortunately, their exact plans are not known  even in a MDP environment. This is due to the fact that  either there is no sufficient domain knowledge that is required to manually build a precise plan or there is no exact learning techniques to automate the plan learning in a complicated problem domain. Consequently, the plan uncertainty challenges intention recognition for multiple agents in this research. 

We tackle intention recognition issues through modelling and reasoning with agents' behaviour. First we propose a new representation to model a single agent's behaviour. The plan model contains a set of a sequence of states and actions that prescribe the agent's intention of achieving the goal. It differs from the previous work on using Bayesian network~(BN) to model a single agent's intention~\cite{heinze2004modelling, pereira2010anytime}. A BN model is a descriptive representation and demands significant effort to build the model. Meanwhile, there has not a clear way to compare the BN model for the purpose of conducting intention recognition for multiple agents. Subsequently, we refine the new model by extracting informative features from agents' behaviour and introducing landmarks into the model. The landmarks are representative states that signal possible achievement for agents in the middle stage of completing their tasks. The refinement incorporates a limited set of agents' actions in intention recognition.  
	
Second we adapt a clustering algorithm for  grouping  intention of multiple agents. The algorithm compares their intention models by measuring similarity of probability distributions over potential landmarks and grouping the models to identify their common intentions. Finally, we demonstrate the multi-agent intention recognition techniques in a commonly used problem domain and apply it in a simulation platform for designing a virtual healthy building.  

Our work is the first study of extending single agent recognition into multi-agent intention recognition. The challenge lies in a formal, operational model of intention recognition as well as an efficient solution to clustering the intentions of multiple agents.  We summarise our contributions below.

\begin{itemize}
	\item We provide a new intention model for a single agent that prescribes its behaviours in an uncertain environment;
	\item We further refine the model and propose a way to build the model for the intention recognition purpose;
	\item We adapt a clustering algorithm to group multi-agent intentions; 
	\item We conduct empirical study of the proposed methods over two problem domains. 
\end{itemize}

We organise this article as follows. In Section~\ref{sec:model} and \ref{subsec:bt}, we propose a new intention model that describes how an agent acts in an uncertain environment. We further refine the model through a landmark concept and propose a clustering algorithm to identify joint intentions of multiple agents in Section~\ref{sec:clustering}. We demonstrate the performance of the recognition techniques in a set of experiments in Section~\ref{sec:results}. In Section~\ref{sec:review}, we review the relate works on intention recognition. Finally, we conclude this article and discuss future work.

\section{Intention Model for Intelligent Agents}
\label{sec:model}
We consider a setting of Markov decision process~(MDP)~\cite{bellman1957markovian} when an agent operates in a fully observable environment. To design the agent model, we use the well-known belief-desire-intention~(\emph{BDI}) architecture where {\em belief} represents the agent's perception over the environment, {\em desire} specifies a goal for the agent and {\em intention} is reflected in a series of actions that the agent executes to complete a task~\cite{georgeff1998belief}. 

In the MDP setting, the agent can observe environmental states are denoted by a set of states $S$ upon which the agent conducts the action $A$ over a limited number of time steps $T$. The interaction between the agent and environment composes the agent's plan and reflects its intention to achieve the goal $G$.  We intend to discover the agent's intention based on the observed plan in the interaction.

\section{Behaviour Tree}
\label{subsec:bt}

A plan sequence prescribes how an agent acts upon the observed states over time, and is formulated below. We elaborate a plan sequence is a sequence of state-and-action pairs. 

\begin{defn} 

$PS_T=\left[ \left( { s }_{ { 0 } },{ a }_{ { 0 } } \right) ,\left( { s }_{ { 1 } },{ a }_{ { 1 } } \right) ,\dots ,\left( { s }_{ { T-1 } }, { a}_{ { T-1 }, } \right)| (g, null)  \right]$, where $a \in A$ is an action option selected by the agent from the acton set $A$, $s \in S$ is the environmental states and it starts with the initial state $s_0$ and lasts for $T$ time steps, which ends with either a final goal $g$ or a $null$ state if the plan sequence fails to meet a goal. 
\end{defn}

Due to the stochastic behaviour~(e.g. the probabilistic transition of the environmental states upon the agent's actions) in an uncertain environment, the agent may go through different plan sequences according to the state transition.  To complete a task, the agent has a full plan that contains a number of plan sequences all of which start from the same initial states and may lead to a set of goals or nowhere if the plan fails to meet any goal. 

\begin{figure}[t]
	\centering
	\includegraphics[width=6cm,height=6cm ]{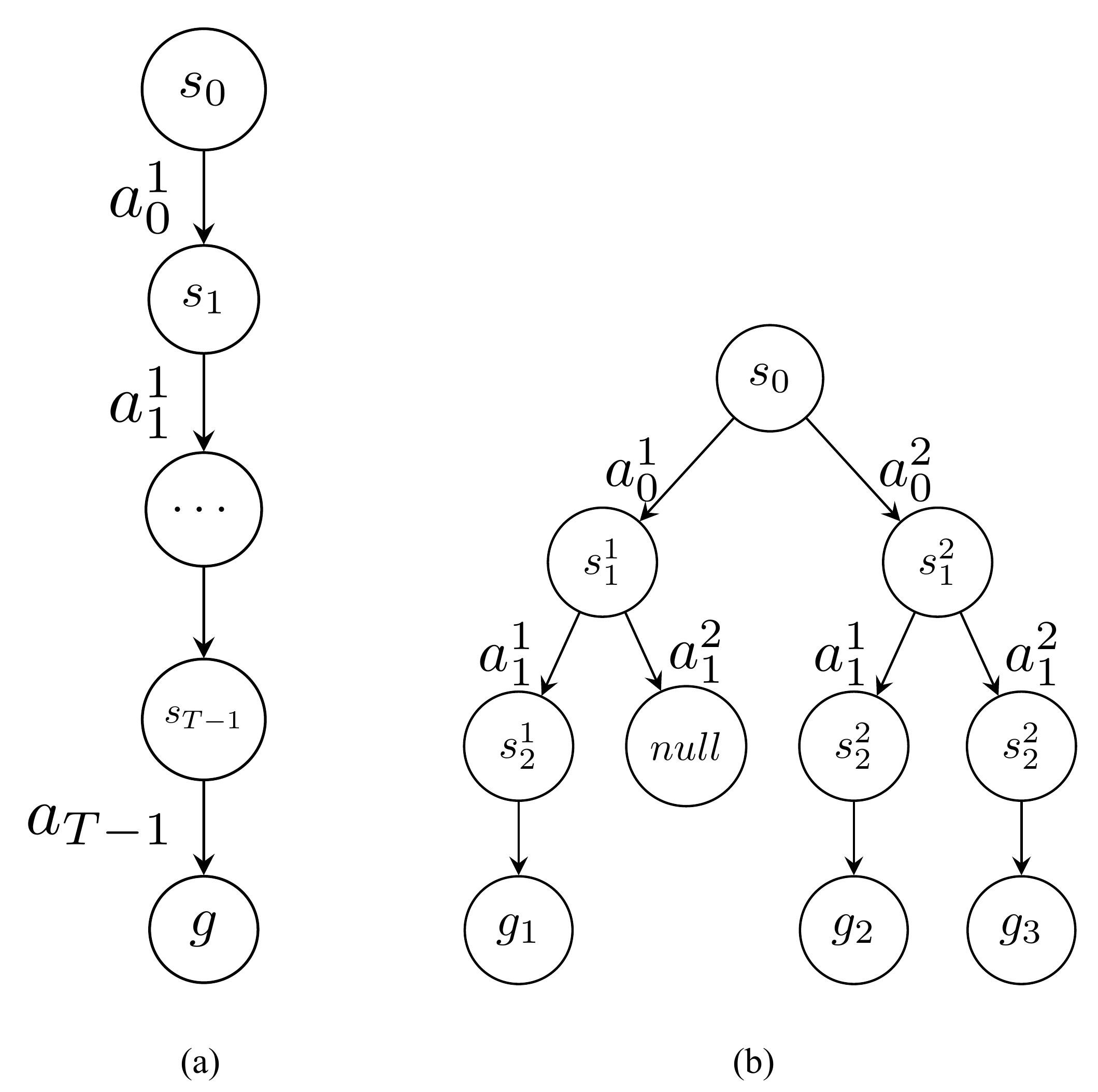}
	\caption{\label{fig:a} $T$-depth behaviour tree
	\label{fig:b} Behaviour Tree with Plan Sequences}
\end{figure}

Formally we can model a full plan of $T$ time steps with the agent's goal $G$ as a $T$-depth behaviour tree $\mathcal{T} = \{\cup PS, (G, Null), T\}$ where a root node is an initial state $s_0$, an internal node represents an environmental state, an arc connecting two states is an action $a$ and the leaf node is a goal state for completing a task.  Fig.~\ref{fig:b} shows a behaviour tree containing a number of plan sequences in Fig.~\ref{fig:a}.

A full plan\cite{doi:10.2514} contains a set of intentions that drive an agent to perform a sequence of actions so as to move towards a final goal. As the intentions are contextualised in the plan, we could infer whether multiple agents share common intentions by comparing the observed plan sequences over time. This provides a cornerstone for studying intention recognition for multiple agents. However, we shall notice that only a partial plan is observed as the agent acts in a real-time setting. This is the only information we could use for intention recognition.

\subsection{Landmark-based Intentions Models}
A behaviour tree prescribes how an agent executes a sequence of actions,  which indicates an intention type that the agent follows to achieve a certain goal. By literally comparing their plans including states and actions, we may discover whether the agents have a common intention. However, the comparison would be rather inefficient, and actually is not necessary and even may compromise the identification of common intentions for multiple agents. A little deviation from a state in a plan execution is not necessary to indicate the difference of their intentions. Hence, we proceed to refine a behaviour tree to develop an intention model that serves for the purpose of grouping intentions for multiple agents. 

Notice that actions are determined by what the agents perceive from the states and what they prefer to act in order to attain a goal. In addition, actions can be directly observed when the agents behave in the environment. Hence, actions are important features for identifying the agents' intentions. Meanwhile, inspired by the fact that a state in an MDP setting is sufficient statistic of state and action sequence in the past, we consider special states, namely {\em landmarks},  in an intention model. A landmark state is more informative than a general state and indicates a clear intention tendency while the agent acts and is observed by the outsider. For example, a waypoint is a useful landmark when a robot navigates in a maze.\cite{10.1007} It indicates that the robot switches into a new zone or has arrived a sub-goal in completing a task. 

By considering action sequences and landmarks in an environment, we propose the following intention model.

\begin{defn} A landmark-based intention model is defined as

$LBIM=\left[  \cup (PS - S), \mathcal{L}, (G, Null), T \right]$

where $\cup (PS - S)$ extracts action sequences from a full plan and is interleaved with a set of landmark states $\mathcal{L}$.  
\label{def:im}
\end{defn}

\begin{figure*}[htbp]
	\centering
	\includegraphics[width=12cm,height=7cm]{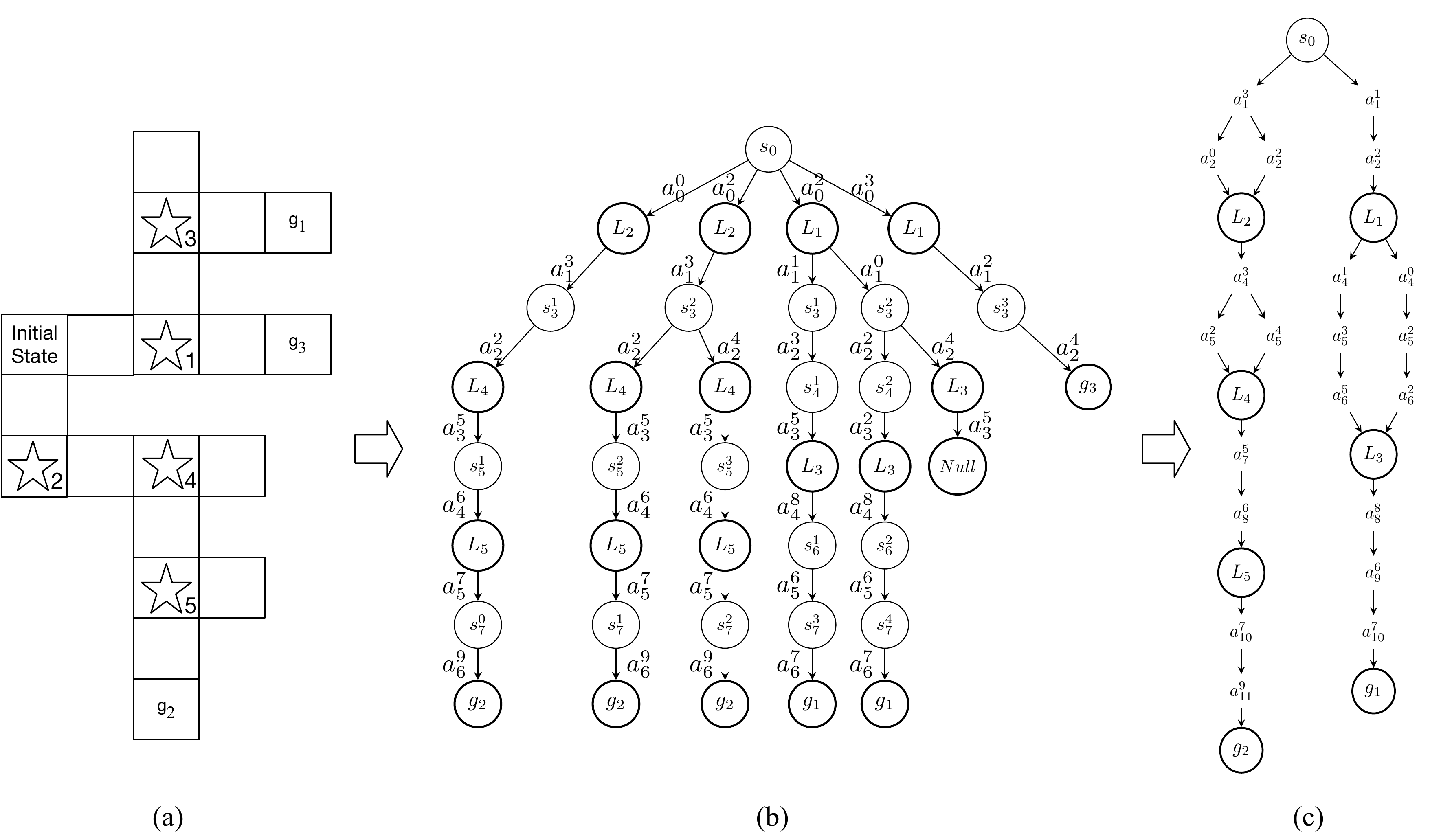}
	\caption{\label{fig:landmark1} An example of the LBIM model.}
\end{figure*}

Fig.~\ref{fig:landmark1} shows one example of a landmark-based intention model~(with all states). In Fig.~\ref{fig:landmark1}~($a$), there are five landmarks~(denoted by a $star$) specified in the hallway and an agent may start from an initial state and gets to the goal~($g_1$--$g_3$) or even stick into the $Null$ state. Fig.~\ref{fig:landmark1}~($b$) is the corresponding LBIM model that contains a set of plan sequences for the agent's navigation. 

A landmark-based intention model facilitates the understanding of an agent's intention based on their actions and distinct states in an environment. Labelling the landmarks from the environment mainly relies on task or domain knowledge in a specific application. The set of landmarks could be a set of (sub)-goals or significant states occurring to the agent, which does not pose much challenge to its application upon a well-defined planning problem. 

Note that the concept of landmarks was used in the traditional AI planning method~\cite{hoffmann2004ordered} and recently is studied in learning policy tree for multiple agents~\cite{conroy2015learning}.  The research also shows benefit of landmarks in understanding agents plan; however, automating a landmark specification is still an open issue in the agent and planning community. 

\begin{figure*}[t]
	\centering
	\includegraphics[width=12cm,height=10cm ]{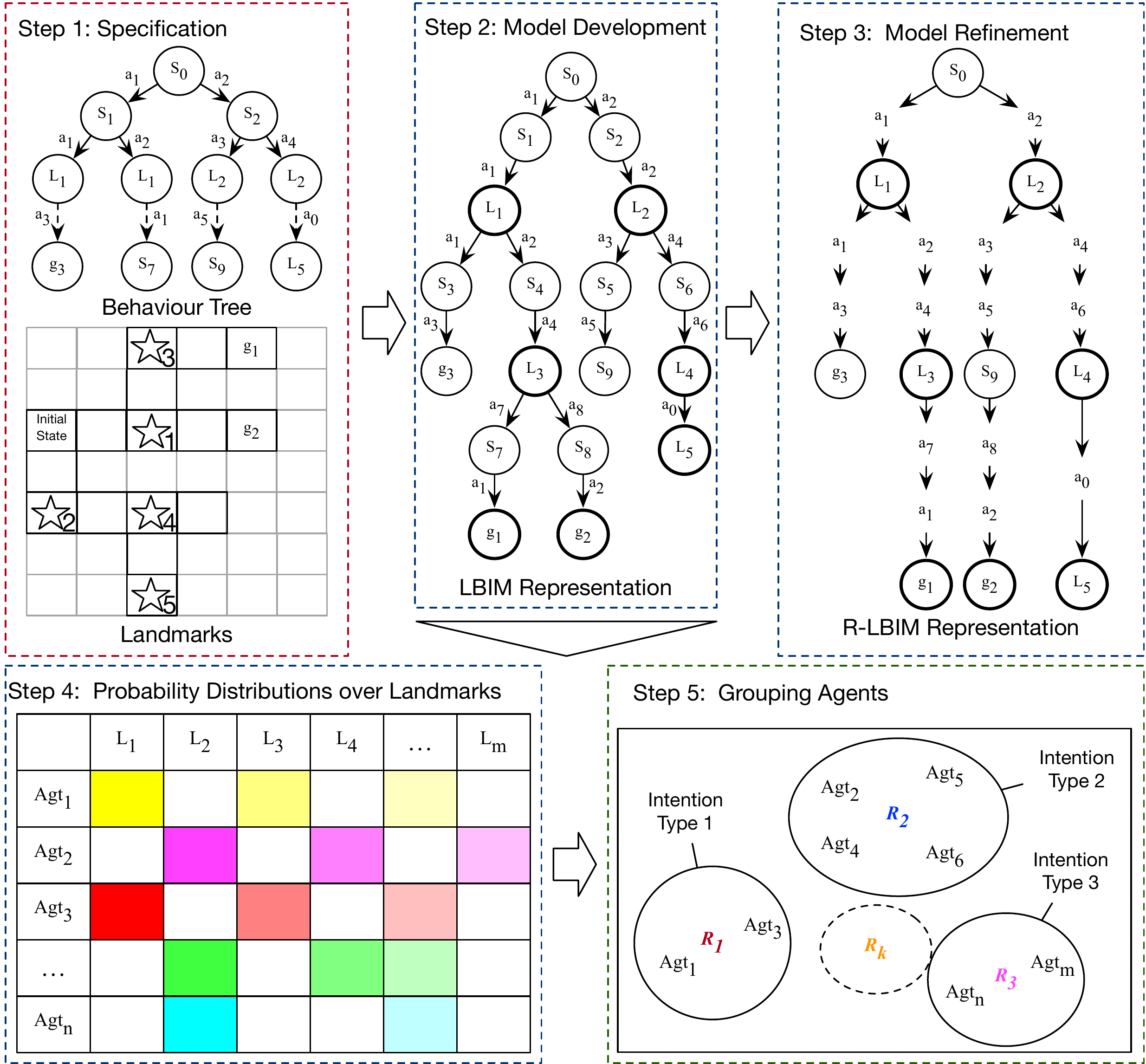}
	\caption{\label{fig:whole} Process of Identifying Common Intentions for Multiple Agents}
\end{figure*}

\section{Grouping Intention Models}
\label{sec:clustering}
Given a definition of intention models, we proceed to solve the problem of identifying a common intention of multiple agents while they act in an uncertain environment. The common intention recognition facilitates their collaboration when the agents are to achieve their goals. The recognition shall be conducted while the agents are executing plans in an online setting. This is challenging since only a partial plan, e.g. a sequence of state and actions in the past, is available for deciding whether they have a common intention in the real-time plan execution.  

More specifically, given the observed state and action sequence of $N$ agents over $t$~($<T$) time steps, denoted by $\mathcal{PS}_t=\left[PS_{1}, \cdots, PS_{N}\right]$, we aim to identify a set of their common intentions by grouping their behaviours,  each of which represents a certain type of a common intention for the agents in the group. Fig.~\ref{fig:whole} shows the main process for the common intention recognition for multiple agents. 

Given the input of their plan sequences that are organised into behaviour trees, we build a landmark based intention model~(LBIM) by considering potential landmarks in a problem domain~(Step 1-2). Subsequently, we refine the LBIM model by extracting common action sequences from the model~(Step 3) based on which we calculate a probability distribution of where the agents are in the corresponding landmarks~(Step 4). Finally, we cluster the agents into several groups in terms of their distributions over the landmarks and each group indicates a potential type of common intentions for the agents within the group~(Step 5). We will elaborate on each step below. 

\subsection{Model Refinement: R-LBIM}
By observing how a group of agents act in an uncertain environment, we can construct intention models for the agents who follow their plan in the mind. As their intentions are hidden in the plan execution therefore driving the agents towards the goal, a landmark is an important state that signals whether the agents are either leading to (sub-)goals or deviating from the goals. 
For example, Fig.~\ref{fig:landmark1}($a$) shows several landmarks, denoted by stars, in the hallway where the agent aims to get to the goals~($g_1$ - $g_3$) by navigating around the grids. The landmark in the bottom~($star_5$) clearly shows that the agent has the intention of getting to the goal $g_2$.

Intuitively an agent exhibits a certain type of intention when it has arrived at a landmark in a plan. The sequence of actions executed by the agent would be different since the agent may take a different plan in completing a task.   The sequence of actions could differ in either the number time steps or an actual action taken in a time step when the agent gets to the landmark. To have a good coverage of how the agent acts upon the landmark arrival, we consider a sequence of common actions possibly taken by the agent over the past.  In other words, we extract a set of common actions over the longest sequence to the landmark. The common actions are representative and shared by the different plans leading to the landmarks, which indicates the hidden intention of the agent. We proceed to elaborate the process of embedding landmarks and extracting the longest common action sequences to refine a landmark-based intention model. 

Given a full plan of an agent~(as represented by a behaviour tree $\mathcal{T}$) and a set of landmarks $L$ specified in a problem domain, we can, in the first phase, replace the corresponding states with the landmarks while leaving out the rest states and actions in the behaviour tree. Table~\ref{table: Landmark Order for Goals} shows that we find the landmarks in the sequences leading to the goals $g_1$ and $g_2$ given the landmark specification in Fig.~\ref{fig:landmark1}~($a$).  Then we can obtain a behaviour tree with the initial states $s_0$, the landmarks and the goals, as shown  in Fig.~\ref{fig:landmark1}~($b$).

\begin{table}[h]
\small\sf\centering
\caption{Landmarks for the Two Goals \label{table: Landmark Order for Goals}}
\begin{tabular}{|l|c|c|c|}
\hline
\diagbox{${ G }_{ n }$}{$Time$} & ${ t }_{ 0 }\rightarrow { t }_{ i }$ & ${ t }_{ i+1 }\rightarrow { t }_{ m }$ & ${ t }_{ m+1 }\rightarrow { t }_{ k }$ \\
\hline
\texttt{${ g }_{ 1 }$} & $0$ & ${L}_{1}$ & ${L}_{3}$ \\
\hline
\texttt{${ g }_{ 2 }$} & ${L}_{2}$ & ${L}_{4}$ & ${L}_{5}$ \\   
\hline
\end{tabular}
\end{table}

Once we get a skeleton of a behaviour tree with an initial state and specified landmarks, we need to extract the longest common action sequences from a full plan. We describe the algorithm for extracting the sequences from the initial state $s_0$ to the landmark  $L_c$ in Algorithm 1. 

The algorithm starts with extracting all the action sequences between the initial state $s_0$ and the landmark $L_c$~(line 4) and then compares each pair of action sequences in the collected set $Q_c$~(line 5-19).  We use $c[i, j]$ to record the length of the longest common action sequence~(line 4) and compare each action in the selected pair $q_x$ and $q_y$ from the $Q_c$~(line 8). We update the common action sequence as shown below. We finally return all the  longest common action sequences $CAQ$ for all the paths between $s_0$ and any landmark. 
$C\left[ i,j \right] =\begin{cases} 0\left( i=0|j=0 \right)  \\ C\left[ i-1,j-1 \right] +1\left( i,j>0\& { q }_{x }={ q}_{ y } \right)  \\ \max { \left( C\left[ i,j-1 \right] ,C\left[ i-1,j \right]  \right)  } \\\left( i,j>0\& { q }_{ x }\neq { q }_{ y } \right)  \end{cases}$

\begin{algorithm}
\caption{Extract the longest common action sequences}
\label{alg:LCS}
\begin{algorithmic}[1]
\REQUIRE $\mathcal{L}=\left[ { L }_{ 1 },{ L }_{ 2 },{ L }_{ c },\cdots ,{ L }_{ e }\right] $, $\mathcal{T}$
\ENSURE $CAQ$

\STATE $ { s }_{ 0 }\leftarrow$ Extract the initial state from $ \mathcal{T}$

	\FOR{$ { L }_{ c }$ in $\mathcal{L}$}
	\STATE $Q_c=\left[ { q }_{ 1 }, { q }_{ 2 }, \cdots , { q }_{ m }\right] \leftarrow $ Extract all the action sequences between $s_0$ and $L_c$ 
	\STATE $c\left[ i,j\right] \leftarrow 0$
\FOR{$q_x,q_y$ in ${ Q }$}
\FOR{$i\leftarrow 1$ to $q_x$}
	\FOR{$j\leftarrow 1$ to $|q_y|$}
	\IF{${ q }_{ x }\left[ i \right] == { q }_{ y }\left[ j \right]$}
	\STATE $c\left[ i,j \right] \leftarrow c\left[ i-1,j-1 \right] +1$
	\STATE $CAQ_{L_c}\leftarrow { q}_{ x }\left[ i-1 \right]$
	\ELSIF{$c\left[ i-1,j \right] > c\left[ i,j-1 \right]$}
	\STATE $c\left[ i,j \right] \leftarrow c\left[ i-1,j \right]$
	\ELSIF {$c\left[ i-1,j \right] < c\left[ i,j-1 \right]$}
	\STATE $c\left[ i,j \right] \leftarrow c\left[ i,j-1 \right]$
	\ENDIF
		\ENDFOR
\ENDFOR
\ENDFOR
\ENDFOR
\RETURN $CAQ=\cup CAQ_{L_c}$
\end{algorithmic}
\end{algorithm}

We use the longest common action sequences to fill in a skeleton of a behaviour tree accordingly. The resulting model refines the LBIM model in Def.~\ref{def:im} by introducing the longest common action sequences through executing the aforementioned algorithm.  This leads to a LBIM refinement model for intention recognition as defined below. 
 
 \begin{defn} A refined landmark-based intention model~(R-LBIM) is defined as 
$R$-$LBIM$=$\left[CAQ, \mathcal{L}, (G, Null), T \right]$ 
where $CAQ$ are the longest common action sequences  and are joined by a set of landmark states $\mathcal{L}$.  
\label{def:rim}
\end{defn}
 
 Fig.~\ref{fig:landmark1}~($c$) shows a R-LBIM example~(for two goals $g_1$ and $g_2$) given the example in Fig.~\ref{fig:landmark1}~($b$). We build the tree skeleton with the landmarks leading to $g_1$ or $g_2$ and fill in the action sequences through running the algorithm in Algorithm ~\ref{alg:LCS}.

\subsection{Inferring Agents' Intentions}
As we do not know a true intention that an agent holds in implementing the course of actions, we can only infer it from what it has acted in the past. To understand whether multiple agents have a common intention in achieving a common goal, we first need to predict potential intentions for individual agents and then check whether their intentions could form common intentions.

For a single agent, we infer its intentions by estimating how it acts towards or has already in a set of landmarks.  Given the observed actions, denoted by $ObsAQ_i=\left[ a_0, a_1, \cdots, a_t \right]$ for the agent $i$, we calculate the similarity $Sim(ObsAQ_i$ $,CAQ_{L_m})$ between the observed action sequence and the common action sequence $CAQ_{L_m}$ leading to every landmark $L_m$ in an intention model. The sequence $CAQ_{L_m}$ is an action sequence that starts from the initial state $s_0$ to ends with the landmark $L_m$ in a R-LBIM model.  The similarity computes the percentage of the common actions overlapping between $ObsAQ_i$ and $CAQ_{L_m}$, and is an indication of how likely the agent's behaviour is linked to the landmark $L_m$. Hence, we can get the likelihood for the agent $Agt_i$ being over the landmarks, $\mathcal{L}=\left[L_1, \cdots, L_M\right]$, and normalise the likelihood values to get the probability distribution over the set of landmarks

$\\Pr(Agt_{ i }|{ L })=\left[ \begin{matrix} \frac { Sim(ObsAQ_{ i },CAQ_{ L_{ 1 } }) }{ \sum _{ m } Sim(ObsAQ_{ i },CAQ_{ L_{ m } }) } , \\ \cdots ,\frac { Sim(ObsAQ_{ i },CAQ_{ L_{ m } }) }{ \sum _{ m } Sim(ObsAQ_{ i },CAQ_{ L_{ m } }) } , \\ \cdots ,\frac { Sim(ObsAQ_{ i },CAQ_{ L_{ M } }) }{ \sum _{ m } Sim(ObsAQ_{ i },CAQ_{ L_{ M } }) }  \end{matrix} \right]$

We can calculate the probability distribution over the landmarks for $N$ agents. Given the probabilities of $N$ agents over the landmarks\\ $\left[Pr(Agt_1|\mathcal{L}), \cdots, Pr(Agt_i|\mathcal{L}), \cdots, Pr(Agt_N|\mathcal{L})\right]$, we proceed to infer their common intentions till the time step $t$.

We adapt the commonly used clustering method, namely $K$-means~\cite{ogston2003method}, to cluster $N$ agents based on their probability distributions over the landmarks. $K$-means is a method of clustering samples according to their probability distribution. We present the algorithm in Fig.~\ref{alg:Clustering}\cite{sinaga2020unsupervised}\cite{mesbahi2015multi}.

We use the \emph{ Kullback®Leibler divergence}~\cite{coma2003multi} to calculate the similarity of the probability distributions of agents and set them into several $K$ sets~(line 4).  We have the agents in one group that have the most similar probability distributions over the landmarks~(line 5). The process is repeated until the average divergence between agents is not changed any more~(line 7).

In Alg.~\ref{alg:Clustering}, the  value of $K$ specifies the number of groups of intentions into which we expect to cluster $N$ agents and each group is one type of intention for the agents. Based on the probability distribution of the agents over the landmarks, we can initialise $K$ as the number of the most probable landmarks where the agents are moving towards or have arrived. We also examine the impact of $K$ on the recognition performance in the empirical study. 

\begin{algorithm}
\caption{Clustering $K$ sets agents over landmarks}
\label{alg:Clustering}
\begin{algorithmic}[1]
\REQUIRE $\left[Pr(Agt_1|\mathcal{L}), \cdots, Pr(Agt_i|\mathcal{L}), \cdots, Pr(Agt_N|\mathcal{L})\right]$, $K$
\ENSURE $K$ groups of agents $\left[ R_1, \cdots, R_K\right]$

\FORALL{ Agents in $\left( { Agt }_{ 1 },{ Agt }_{ 2 }, \cdots ,{ Agt}_{ N } \right)$}
\STATE Randomly choose $K$ agents as centres
\REPEAT 
\STATE Calculate the similarity degree between the centred agent $Agt_i$ and other agent $Agt_j$ through the distance, \\${ D }_{ KL }\left( Pr(Agt_{ i }|{ \mathcal{L} })\parallel Pr(Agt_{ j }|{ \mathcal{L} }) \right) =\sum _{ i=1 }^{  }{ Pr(Agt_{ i }|{ \mathcal{L} }) } \log { \frac { Pr(Agt_{ i }|{ \mathcal{L} }) }{ Pr(Agt_{ j }|{ \mathcal{L} }) }  }$
\STATE Put the agent $Agt_j$ into the most similar group $R_i$
\STATE Calculate the new centers ${ Agt }_{ i }=\frac { 1 }{  |{ R }_{ j } |  } \sum _{ i\in { R }_{ j } }^{  }{ { Pr(Agt_{ i }|{ \mathcal{L}}) } } $
\UNTIL {there is no change in the groups}
\ENDFOR

\RETURN $\left[ R_1, \cdots, R_K\right]$\end{algorithmic}
\end{algorithm}

\section{Experimental Results}
\label{sec:results}
We conduct a set of experiments to demonstrate performance of intention recognition techniques in two problem domains. One is the {\em Tileword} problem~\cite{lees2002history} that is well developed in multi-agent research while the other is in one 3D simulation platform that is dedicated to investigate impact of social behaviour in designing a healthy office building~\cite{zeng2017using}. As this is the first work on developing a formal method for recognising common intentions for multiple agents, our objectives of the experiments are to: ($a$) verify benefit of multi-agent intention recognition in facilitating agents' collaboration thereby leading to their better performance; and ($b$)  show the performance of our techniques and discuss the improvement. The experiments will also develop our awareness of practical applications of the research on multi-agent intention recognition. 

\subsection{Experimental Settings: Planning and Intention Recognition}
In each problem domain, we use a reinforcement learning technique to learn a full plan, which is always approximate, for agents, and the agents will execute the plans that are formulated as behaviour trees as defined in Def.~\ref{subsec:bt}.

\begin{figure}[htbp]
	\centering
	\includegraphics[width=6cm,height=4.5cm]{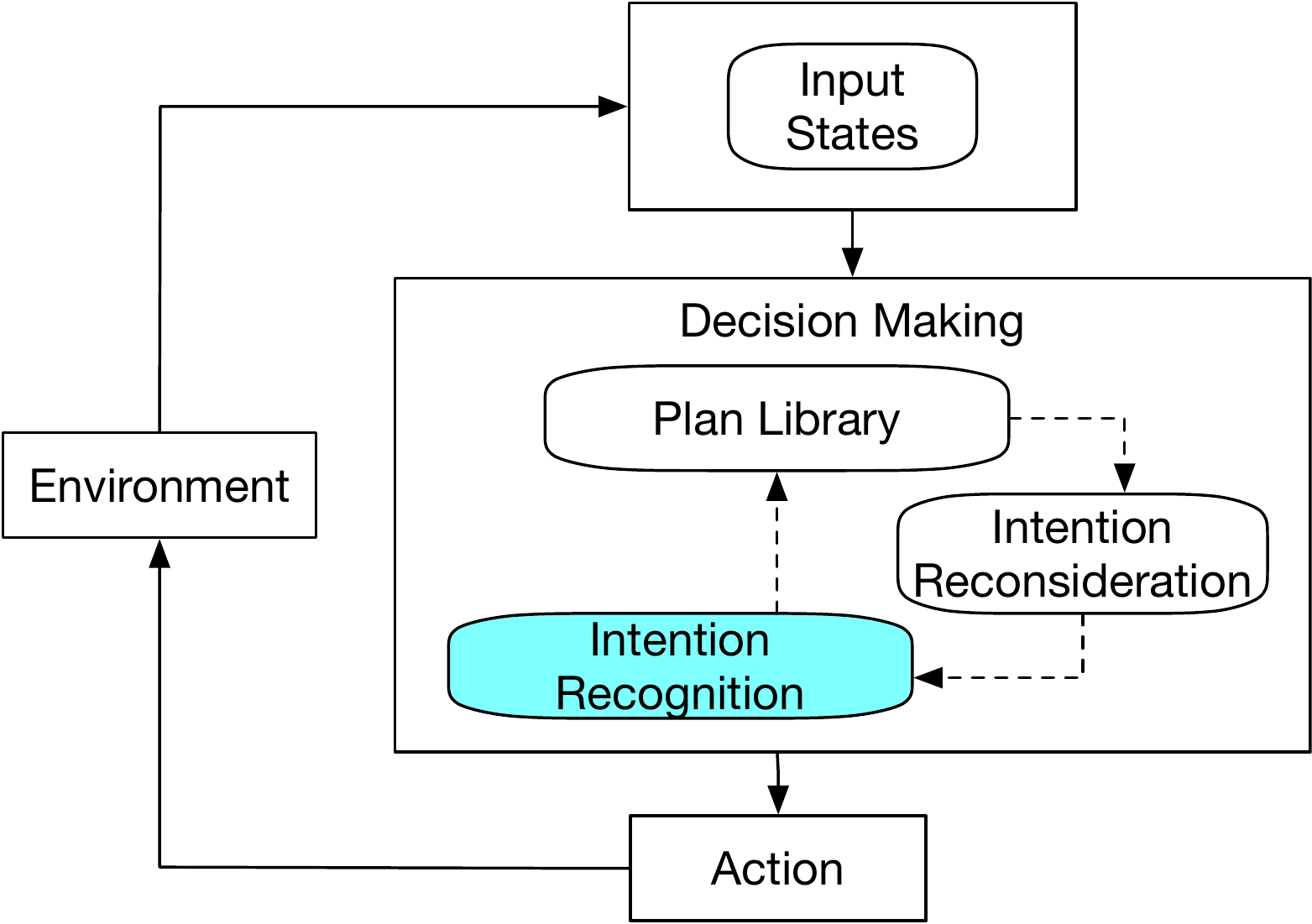}
	\caption{\label{fig:bdi} Belief-Desire-Intention implementation with intention recognition}
\end{figure}

To implement and run the agents in the experiments, we use a BDI architecture and have the plan that commands how the agents shall react given the input of the environmental states. In the BDI implementation, we develop intention reconsideration strategies following a general rule, as suggested by the research~\cite{georgeff1998belief} that the frequency of reconsidering the intentions depends on the dynamics of the problem domain.  Fig.~\ref{fig:bdi} shows the entire process that integrates the BDI architecture with intention recognition function. 

Once the agents decide to reconsider the intentions, they embark  on  our techniques to recognising intentions for multiple agents and identify any possible collaboration with other agents given the discovered common intentions. This is the way how multi-agent intention recognition plays a role in the agent planning for solving various tasks in the problem domains of interest.  

\subsection{Experimental Results}
We implement the multi-agent recognition techniques and the BDI agent architecture. We implement two multi-agent recognition techniques. The first technique, denoted by LBIM, uses the landmark based intention models based on which we group the models directly to identify potential common intentions for multiple agents. The second one, denoted by R-LBIM, is the LBIM refinement by using the longest common action sequences for the clustering purpose. To conduct a comparison with our techniques, we also implement the BN based intention recognition technique~(BNS) that, however, is for a single agent setting.  We compare all the techniques in different scenarios through a number of measurements.

\begin{figure}[htbp]
	\centering
	\includegraphics[width=8cm,height=3.5cm]{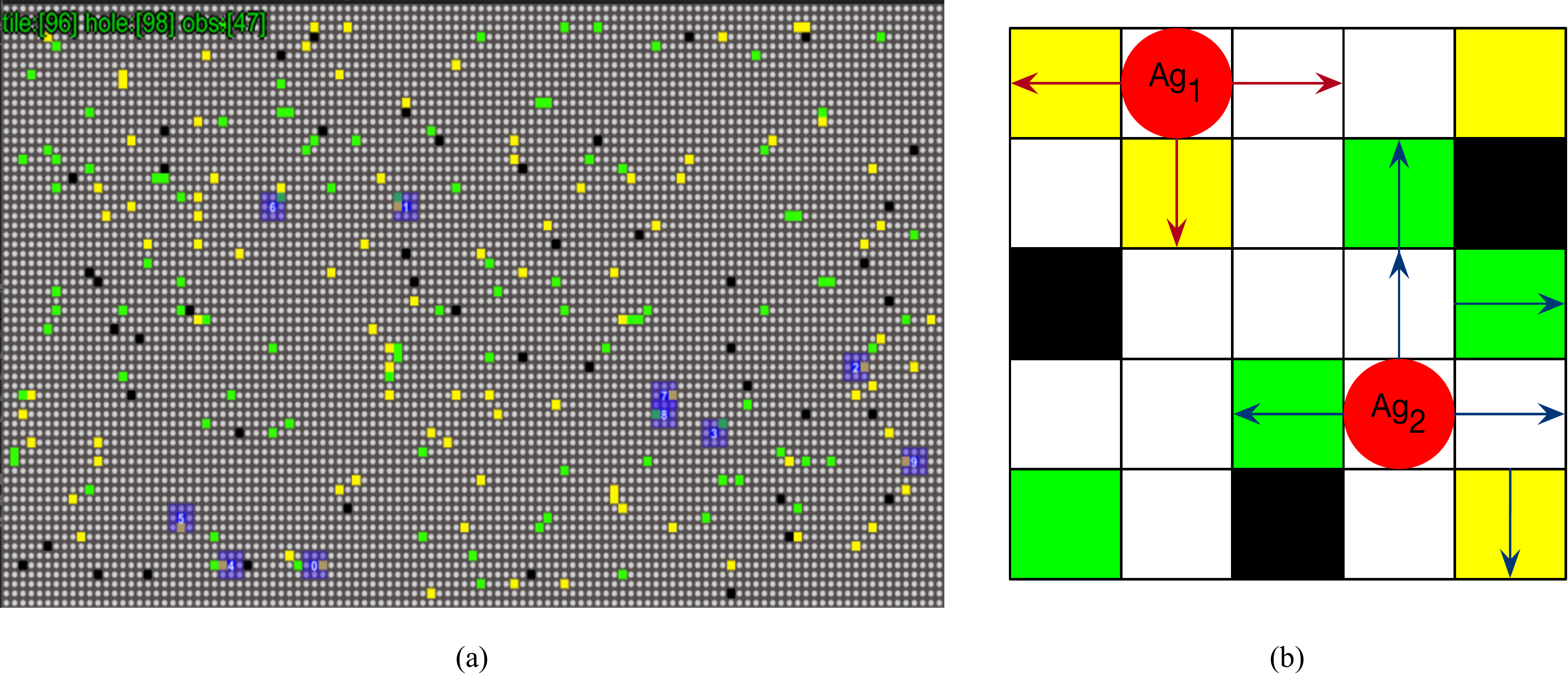}
	\caption{\label{fig:re1} ($a$) A scalable {\em Tileworld} setting
	\label{fig:tw} ($b$) Details of the setting}
\end{figure}

\subsection{The {\em Tileworld} Problem Domain}
Fig.~\ref{fig:tw}($a$) shows the {\em Tileworld} problem domain of 100*100 grids. It is a grid-world consisting of three types of blocks, namely  \emph{Tiles}(Yellow), \emph{Holes}(Green) and \emph{Obstacles}(Blacks). An agent~(Red) has four types of movements - \emph{up}, \emph{down}, \emph{left} and \emph{right}, and needs to avoid the obstacles in the navigation. When the agent gets to a grid, it can either pick up a tile or drop it into a hole if the agent has held the tile. The agent obtains the score of value of \emph{10} when it succeeds in the tile delivery.  Some of the titles in the grids are light and one agent can lift and move it into the hole by itself. However, other tiles demand the effort from at least two agents and they will share the scores upon completing the delivery through their collaboration. To enrich the experiments, we also add a new action that allows the agents to push away the obstacles and they or other agents can pass through the gird, which increases their collaboration chance.

We evaluate the technique performance through two measurements: ($a$) the first one is a $Score$ value that measures a total number of scores achieved by an agent within a certain amount of time; and ($b$) the second one is an $Efficiency$ value that measures how fast an agent can achieve the targeted score at any specific time within a certain period.  We divide an entire period into three time stages, namely the $Early$, $Middle$ and $Final$ stages and compare the performance over  time.  We run the simulations for around 200 times and calculate the averaged performance measurements.

\begin{figure*}[ht]
\begin{minipage}[t]{0.5\linewidth}
\centering
\includegraphics[width=2.0in ]{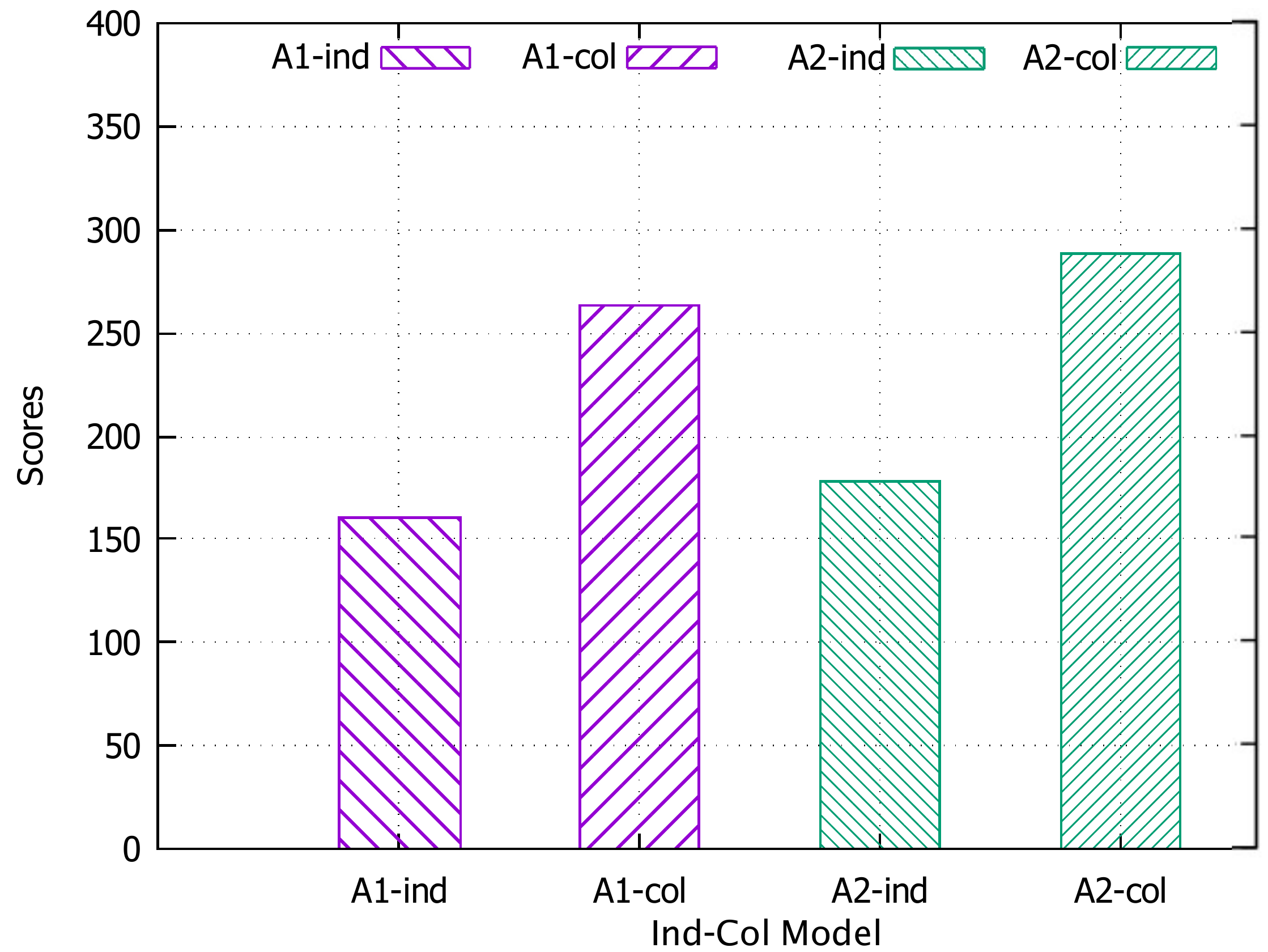}
\\{\small ($a$) Two-agent~($A_1$ -$A_2$) setting}
\end{minipage}%
\begin{minipage}[t]{0.5\linewidth}
\centering
\includegraphics[width=2.0in ]{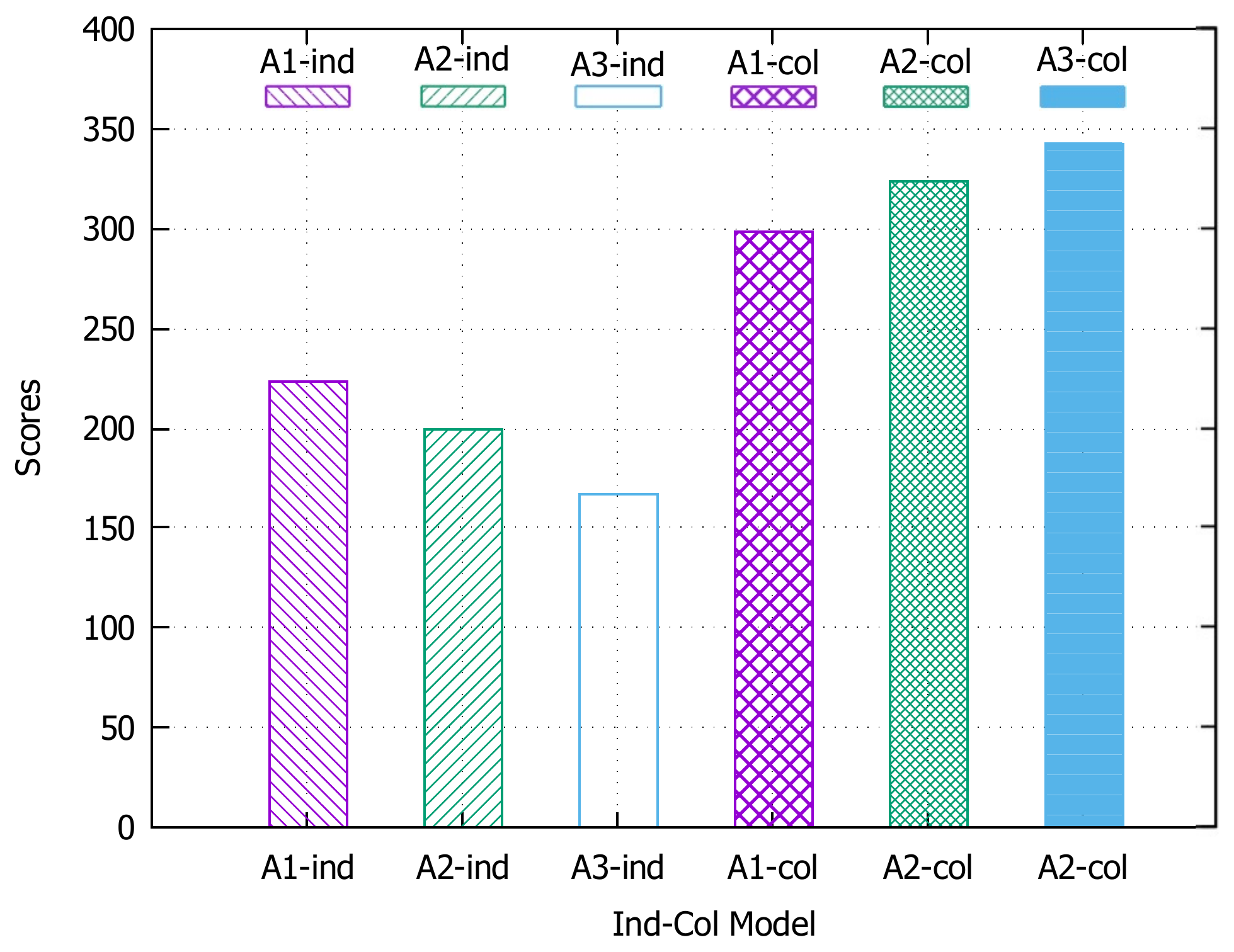}
\\{\small ($b$) Three-agent~($A_1$ -$A_3$) setting}
\end{minipage}
\begin{minipage}[t]{0.5\linewidth}
\centering
\includegraphics[width=2.0in ]{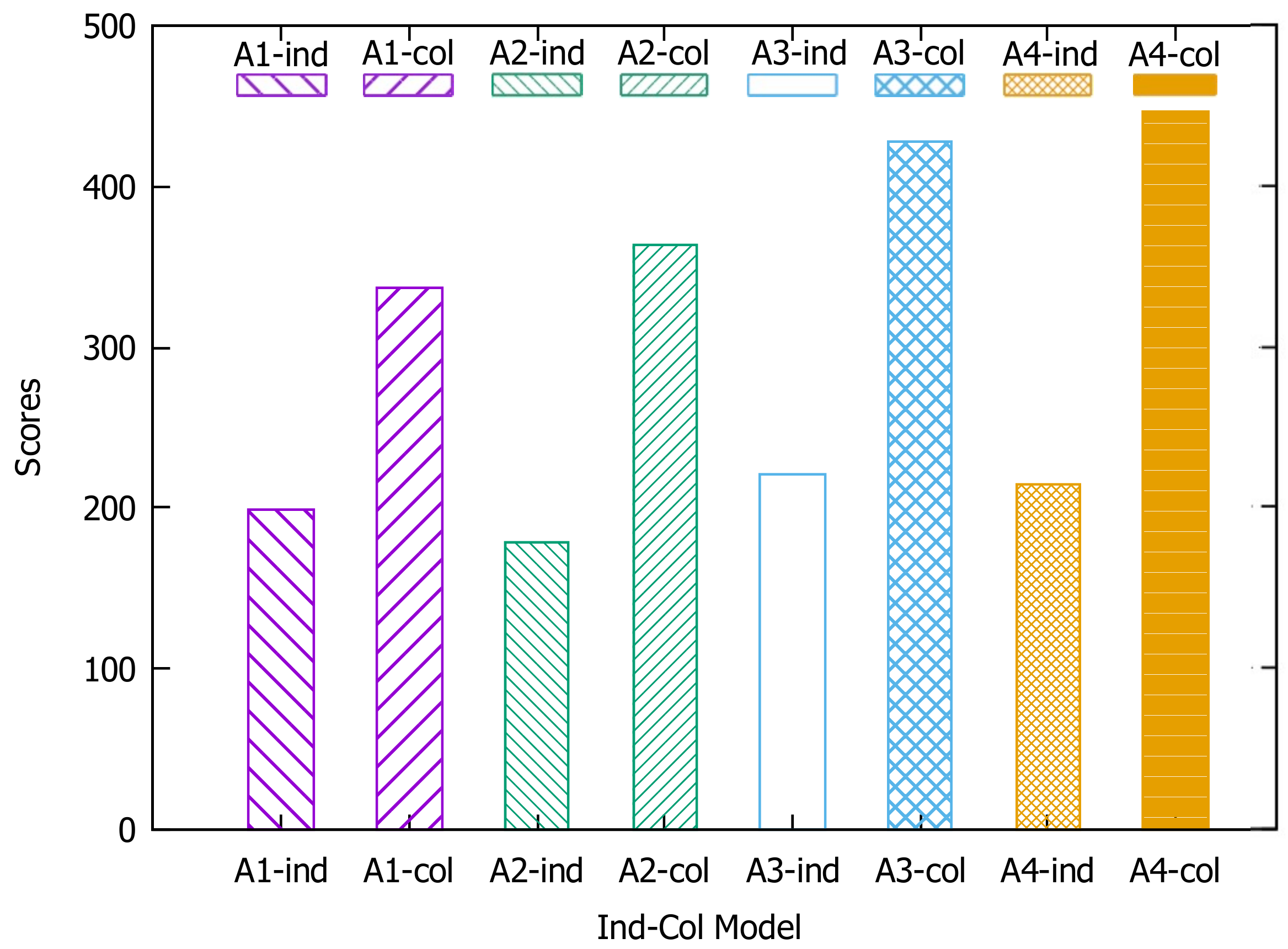}
\\{\small ($c$) Four-agent~($A_1$ -$A_4$) setting}
\end{minipage}
\begin{minipage}[t]{0.5\linewidth}
\centering
\includegraphics[width=2.5in ]{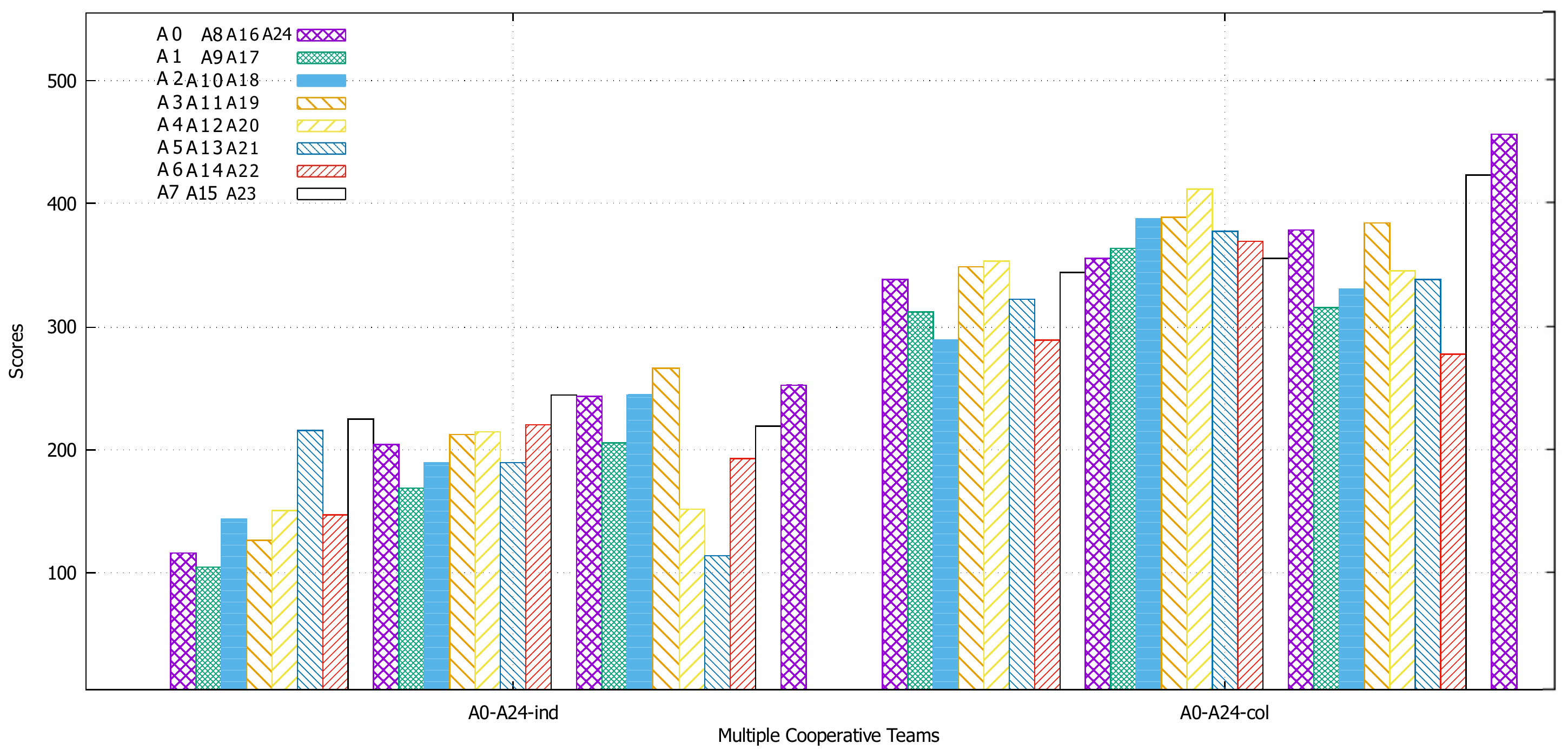}
\\{\small (($d$)Many agents~$\left( up\  to\  25\right)$ setting}
\end{minipage}
\caption{Agents achieve better scores when they collaborate through multi-agent recognition techniques.}
\label{fig:Collaboration}
\end{figure*}

In Fig.~\ref{fig:Collaboration}, we verify the benefits of using multi-agent intention recognition to exploit potential collaborations among multiple agents. We consider two scenarios: the first one lets every agent conduct its own tasks to maximise its scores  while the other asks the agent to identify potential collaboration with other agents when it moves around the grids. Fig.~\ref{fig:Collaboration}~($a$) - ($c$) shows the comparison of the two scenarios in the settings that a small number of agents~($Agt_1 - Agt_4$) aim to achieve a good score within a period of 2 minutes. $A$-$ind$ is the result when the agents work in an individual way while $A$-$col$ arises from the case when the agents use the intention recognition technique of R-LBIM to find out the collaboration.

We increase the number of agents up to 25 in Fig.~\ref{fig:Collaboration}~($d$). We can clearly observe that the agents get better scores when they collaborate in the tile delivery. The benefits are due to the fact that the collaboration provides more opportunities to the agents in completing the tasks. The  intention recognition facilitates their collaboration development in the task completion. 

We proceed to compare our intention cognition technique LBIM with the BNS method in another set of experiments. As BNS is mainly designed for a single agent setting, we adapt it to a multi-agent setting in our implementation. A subject agent first uses BNS to identify intentions of other agents, which involves building a BN model for every agent, and then compares the identified intention with its own intentions to determine the collaboration. We conduct the comparison through different collaboration chances for a large number of the agents~(up to 80). The collaboration scenarios are listed from a simple pair collaboration to a complicated mixed collaboration below. 

\begin{itemize}
	\item FP - an agent can only pair with another agent and the pair of two agents is fixed in the collaboration. In other words, only a fixed pair of agents exist in potential collaboration; 
	\item CP - an agent can pair with different agents and can choose their pairs in the collaboration;
	\item FG - collaboration can occur in a group of more than two agents and the agents are fixed in every group;
	\item MG - an agent can collaborate with a group of different agents in the collaboration;
	\item MM - there is a mixed combination of FG and MG where some agents are free to collaborate with any agents while others can only collaborate with a certain group of agents.
\end{itemize}

\begin{figure*}[ht]
\begin{minipage}[t]{0.5\linewidth}
\centering
\includegraphics[width=2.0in ]{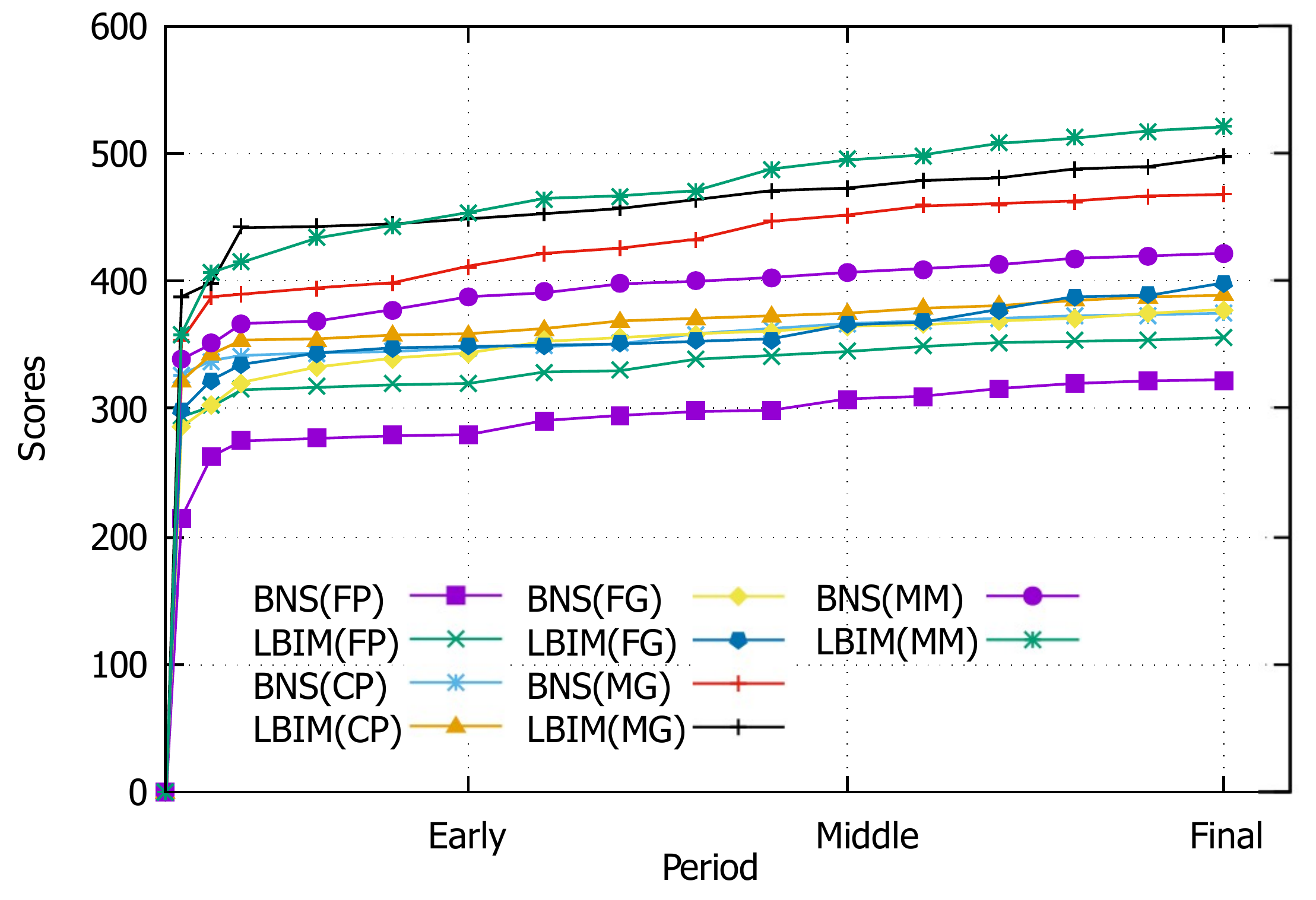}
\\{\small ($a$) Score Evaluation}
\end{minipage}%
\begin{minipage}[t]{0.5\linewidth}
\centering
\includegraphics[width=2.0in ]{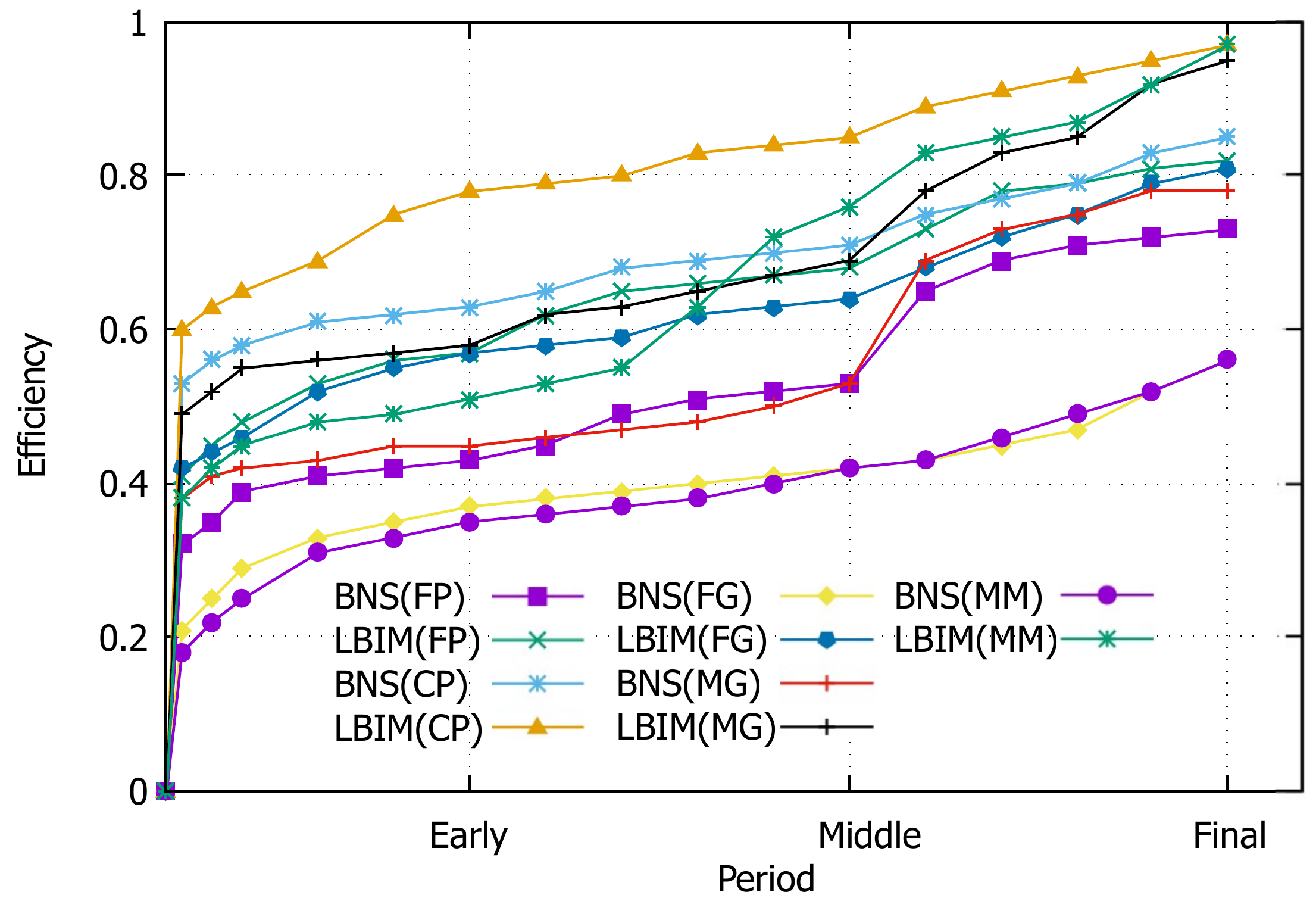}
\\{\small ($b$)Efficiency Evaluation}
\end{minipage}
\caption{LBIM achieves better performance than BNS for different settings.}
\label{fig:BN}
\end{figure*}

Fig.~\ref{fig:BN} shows the score and efficiency of the LBIM and BNS methods. A coloured curve describes the performance resulting from a combination of each method and a collaboration case listed above. Both methods increase the scores and efficiency gradually over the time period. For each collaboration case, LBIM achieves better scores and efficiency than BNS, and most of the gaps are bigger in the final stage. When the simulation approaches the end, a number of tiles that a single agent can lift become fewer and more tasks require collaboration. We observe that BNS(FP) achieves the lowest score while LBIM(MM) has the best outcome. LBIM(CP) is the most efficient since it is free to choose its collaborative pairs.

In a whole, in comparison to LBIM, BNS is a very sophisticated model that is descriptive and is not easily to be built including the parameter estimation and structure learning. We learned the BN based on our simulation data and the resulting model may not be good enough to predict the agents' intention. This compromises the performance of identifying intentions in our simulations.

\begin{figure*}[ht]
\begin{minipage}[t]{0.5\linewidth}
\centering
\includegraphics[width=2.0in ]{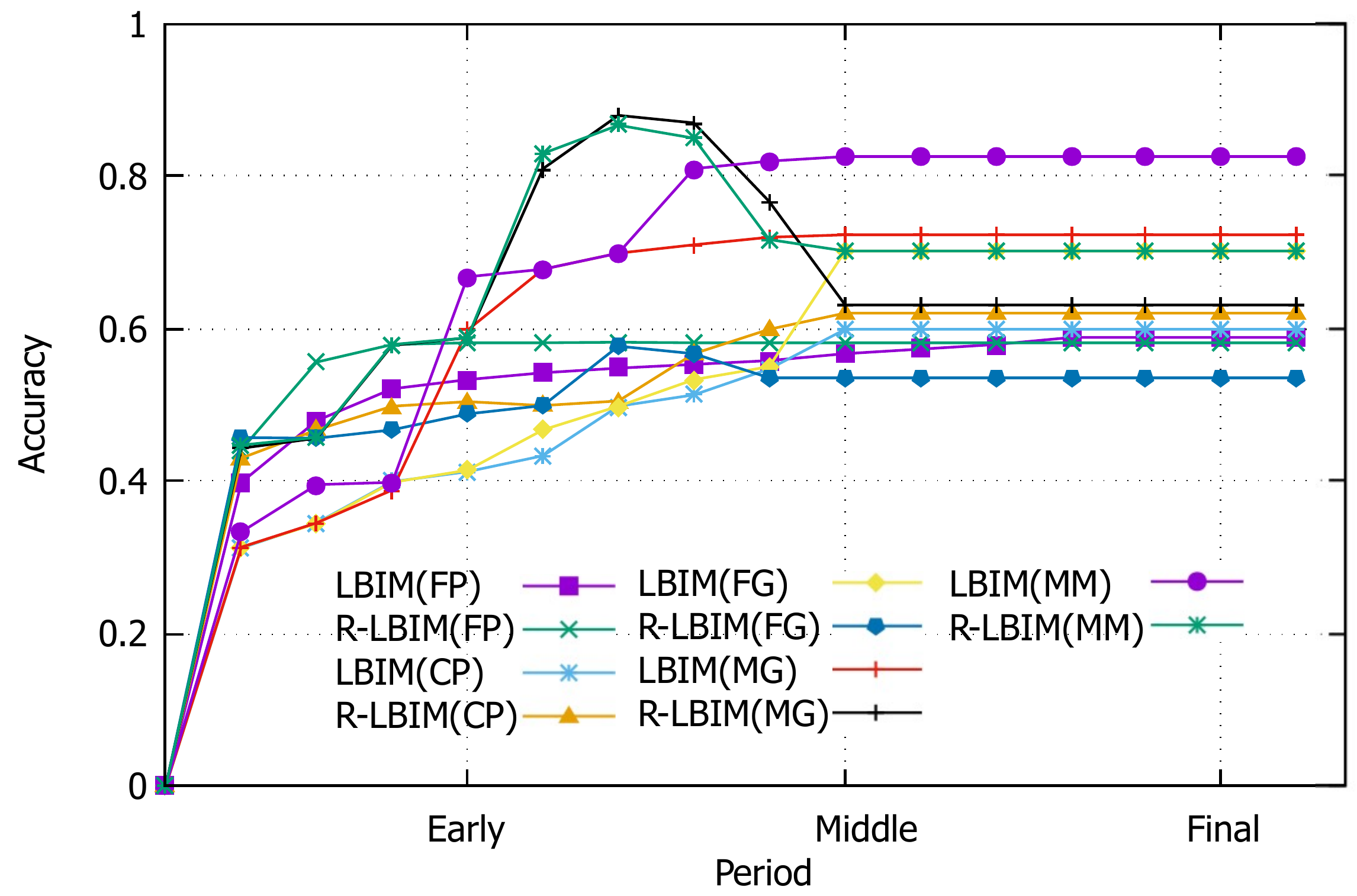}
\\{\small ($a$) Accuracy Evaluation}
\end{minipage}
\begin{minipage}[t]{0.5\linewidth}
\centering
\includegraphics[width=2.0in ]{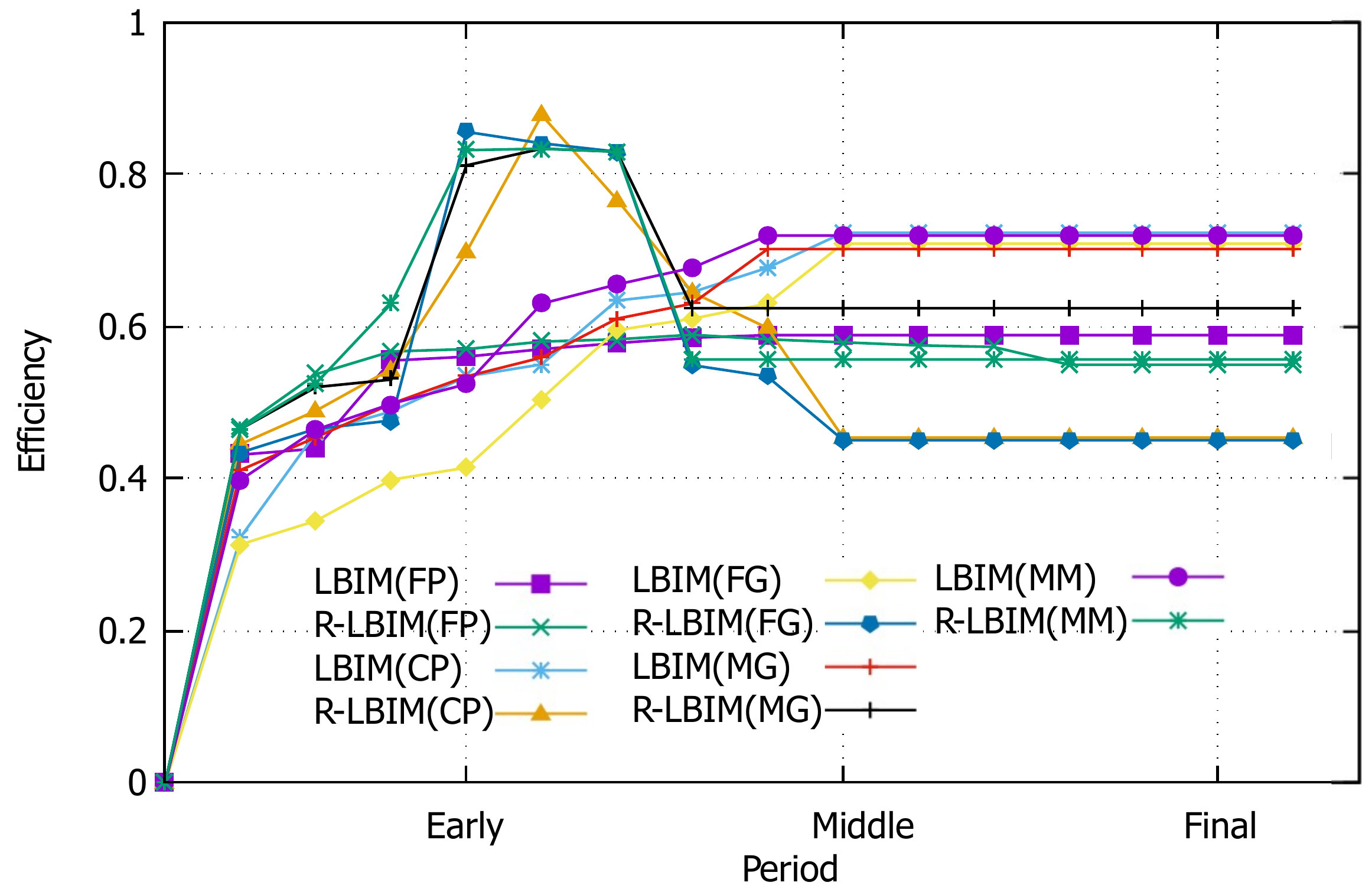}
\\{\small ($b$) Efficiency Evaluation}
\end{minipage}
\caption{Performance of LBIM and R-LBIM given a short length of plan sequences in the intention model.}

\label{fig:LBIM}
\end{figure*}

\begin{figure*}[ht]
\begin{minipage}[t]{0.5\linewidth}
\centering
\includegraphics[width=2.30in ]{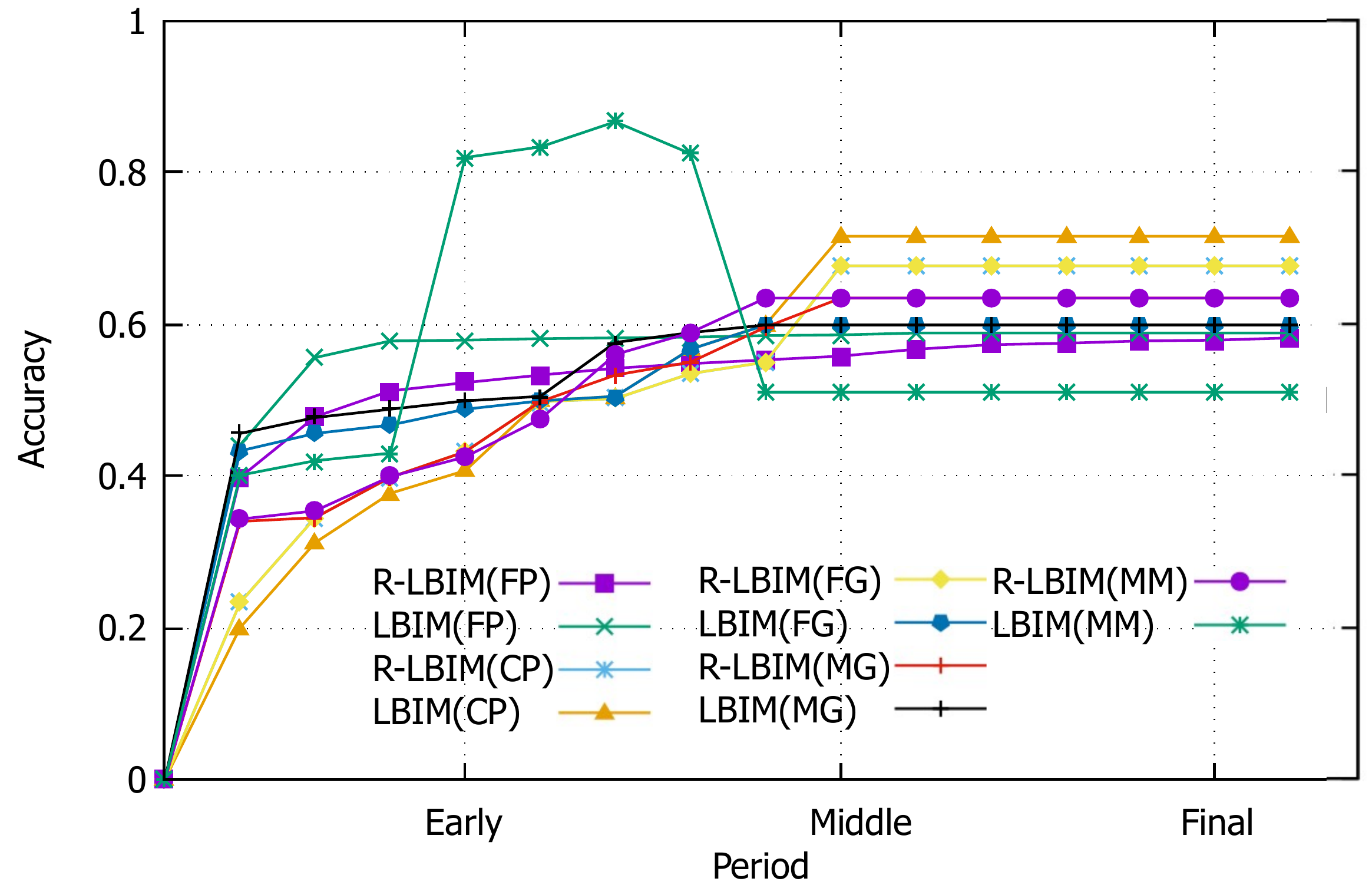}
\\{\small ($a$)Accuracy Evaluation}
\end{minipage}%
\begin{minipage}[t]{0.5\linewidth}
\centering
\includegraphics[width=2.30in ]{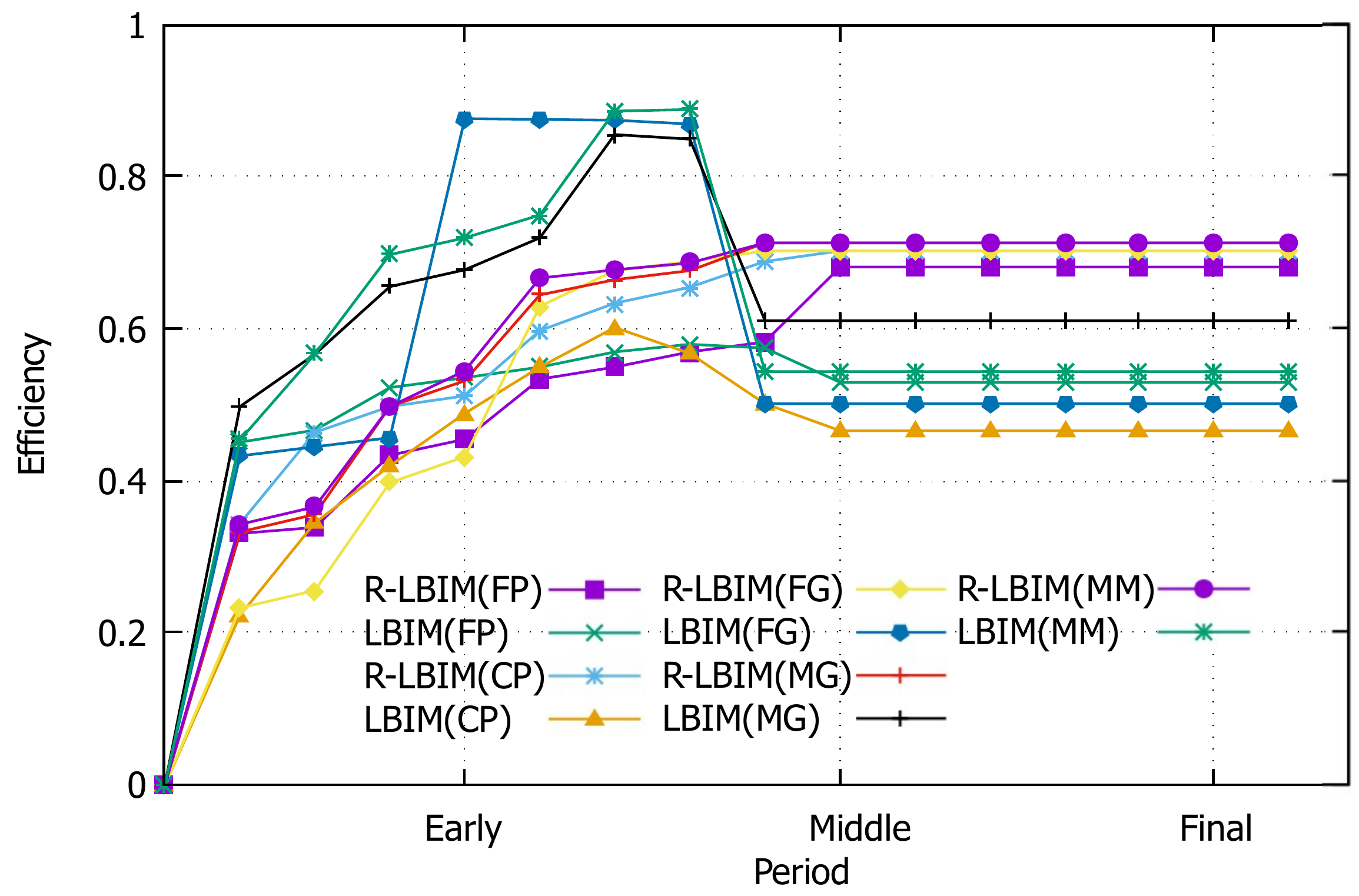}
\\{\small ($b$)Efficiency Evaluation}
\end{minipage}
\caption{Performance of LBIM and R-LBIM given a long length of plan sequences in the intention model.}
\label{fig:LBIM2}
\end{figure*}

Finally we compare the two multi-agent intention recognition techniques, namely LBIM and its refinement version R-LBIM, in the experiments. We consider the comparison in two types of intention models that differ in the length of plan sequences. Fig.~\ref{fig:LBIM} shows both methods achieve similar performance throughout the entire period of conducting tasks. R-LBIM does not achieve better than LBIM when the agents conduct a short sequence of plans. Both methods have a very shallow model of behaviour tree. Both R-LBIM(MG) and R-LBIM(MM) get the best accuracy between the early and middle stages since  a short length of actions can be quickly executed by the agents within a short time and they complete the tasks and enter into the random pair modes. 

In contrast, R-LBIM outperforms LBIM upon executing a long plan sequence in Fig.~\ref{fig:LBIM2}. Notice that LBIM(MM) achieves the best accuracy in the middle stage since its detailed comparison benefits the intention identification when the agents are still running simple behaviour in completing tasks e.g. directly running to lift a titles and dropping it in a hole nearby.  When the behaviour becomes complicated in the final stage, R-LBIM shows better performance. Due to the extraction of common action sequences, R-LBIM is more scalable than LBIM and does not compromise the planning quality in achieving the goals. This becomes more visible in the final stage since the LBIM model becomes more complicated and the refinement reduces more uncertainty of identifying intentions so as to perform better in the simulation. 

\begin{figure}[htbp]
	\centering
	\includegraphics[width=12cm,height=4cm]{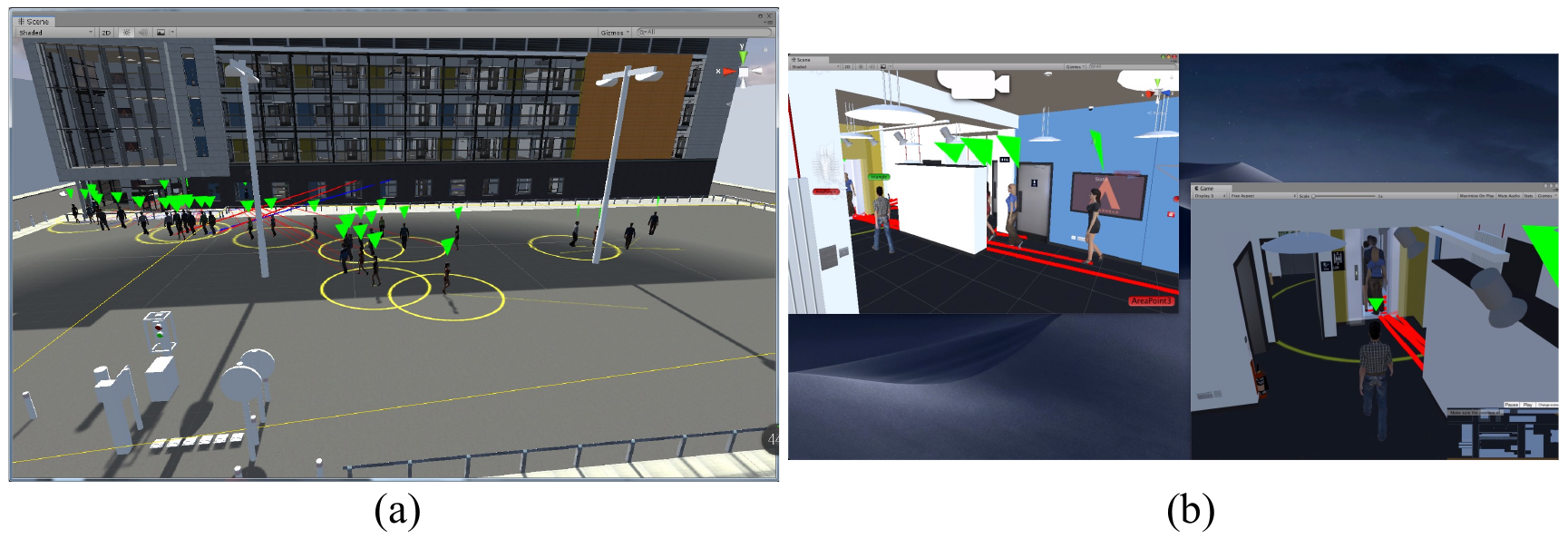}
	\caption{\label{fig:a1} ($a$)~External view of the virtual building
	\label{fig:a2} ($b$)~Crowd moving inside the building}
\end{figure}

\subsection{The Heathy Building Simulation}
We build a 3D simulation platform to evaluate how an office building design will impact physical activities of workforce in the building. This is an extension of our previous development~\cite{zeng2017using} by simulating the existing building~(Phoenix) in the campus of Teesside University. Fig.~\ref{fig:a1}($a$) shows the view of the virtual building from the carpark entrance while Fig.~\ref{fig:a2} ($b$) is a snapshot of the crowd movement in our simulation. One challenging job in the platform development is to simulate human-like behaviour of the crowd when they behave in the building, which leads to more practical evaluations of physical activity level in the building. 

In our experiments, we set a particular task on simulating how people act when they choose to use a lift or walk through a staircase to the first floor. The selection mainly depends on their states including: ($a$) its position relevant to the locations of the lift or the staircase; and ($b$) its perception of other persons' intentions so as to follow their behaviours. The state-based selection mechanism provides  behavioural dynamics in simulating the crowd behaviour. By running a large number of simulations, we can build a set of behavioural models, namely behaviour trees, for individual persons.  The models approximate their navigation plan to the first floor and provide the inputs to our experiments. 

We conduct the experiments in the simulation platform where a crowd of persons enter from the front door and every person reconsiders his intention on his way to the lift or the staircase. We run two sets of experiments to test our intention recognition techniques. In the first set of testing the intention recognition in a single agent setting, we let every person first identify the intentions of each individual of other persons and then decide the next actions by following their common intentions. In the second set of testing the techniques in a multi-agent setting, we, from the third-person perspective, identify common intentions for a number of groups of persons and use the resulting plan to command the movement of the persons. As we observe what the persons are actually moving in the simulation, we can compute the intention recognition {\em Accuracy} by comparing the actual actions of the persons and what our intention recognition techniques predict for them. We also report the {\em Efficiency} of each technique in the experiments and it measures the time~(milli-seconds) consumed to predict common intentions for multiple agents. We compare their performance throughout the simulation period that starts from the door entrance~(the {\em Early} stage), approaches the lift/staircase~(the {\em Middle} stage) and ends with the arrival of the first floor~(the {\em Final} stage). 

\begin{figure*}[ht]
\begin{minipage}[t]{0.5\linewidth}
\centering
\includegraphics[width=2.0in ]{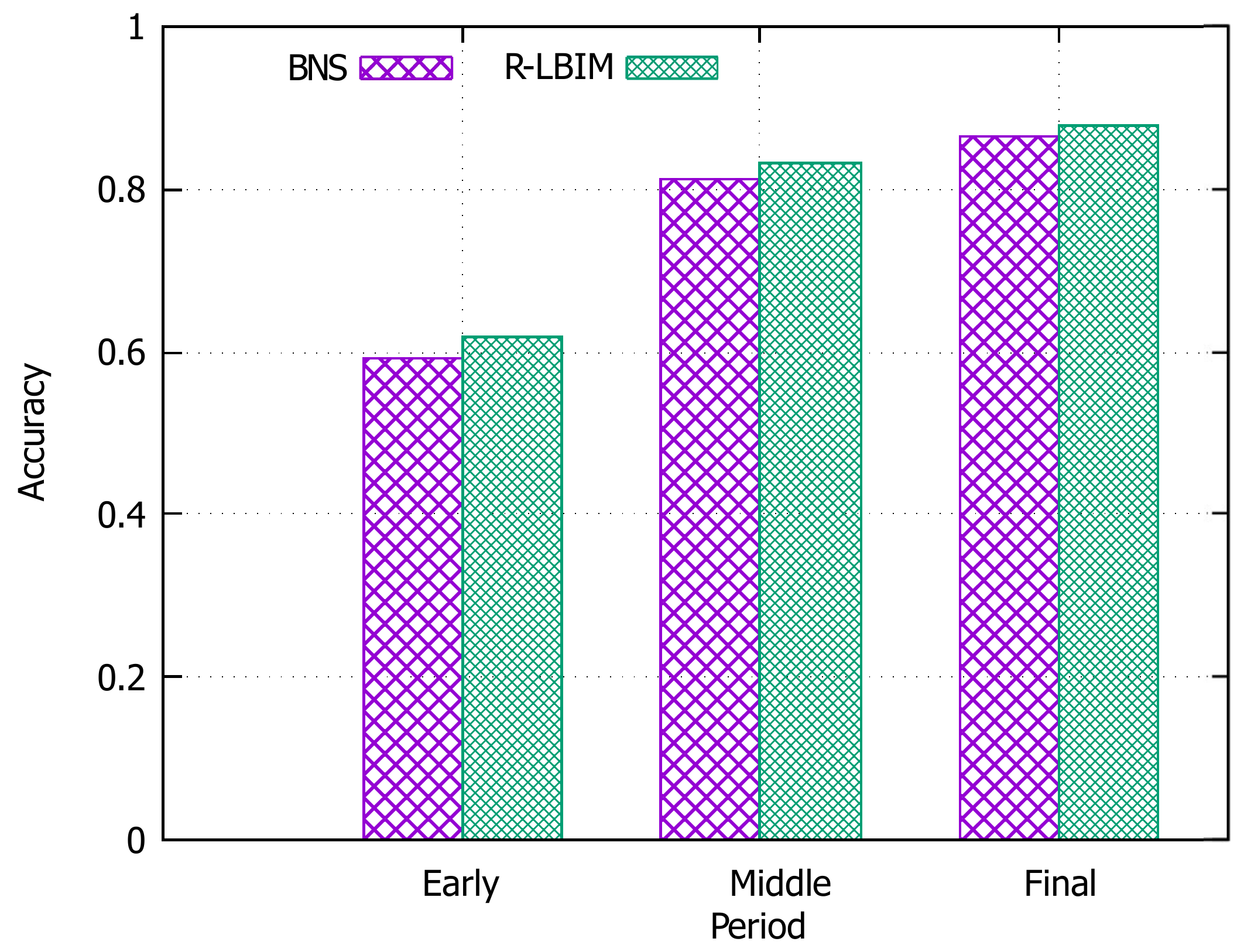}
\\{\small ($a$)~Accuracy Comparison}
\end{minipage}%
\begin{minipage}[t]{0.5\linewidth}
\centering
\includegraphics[width=2.0in ]{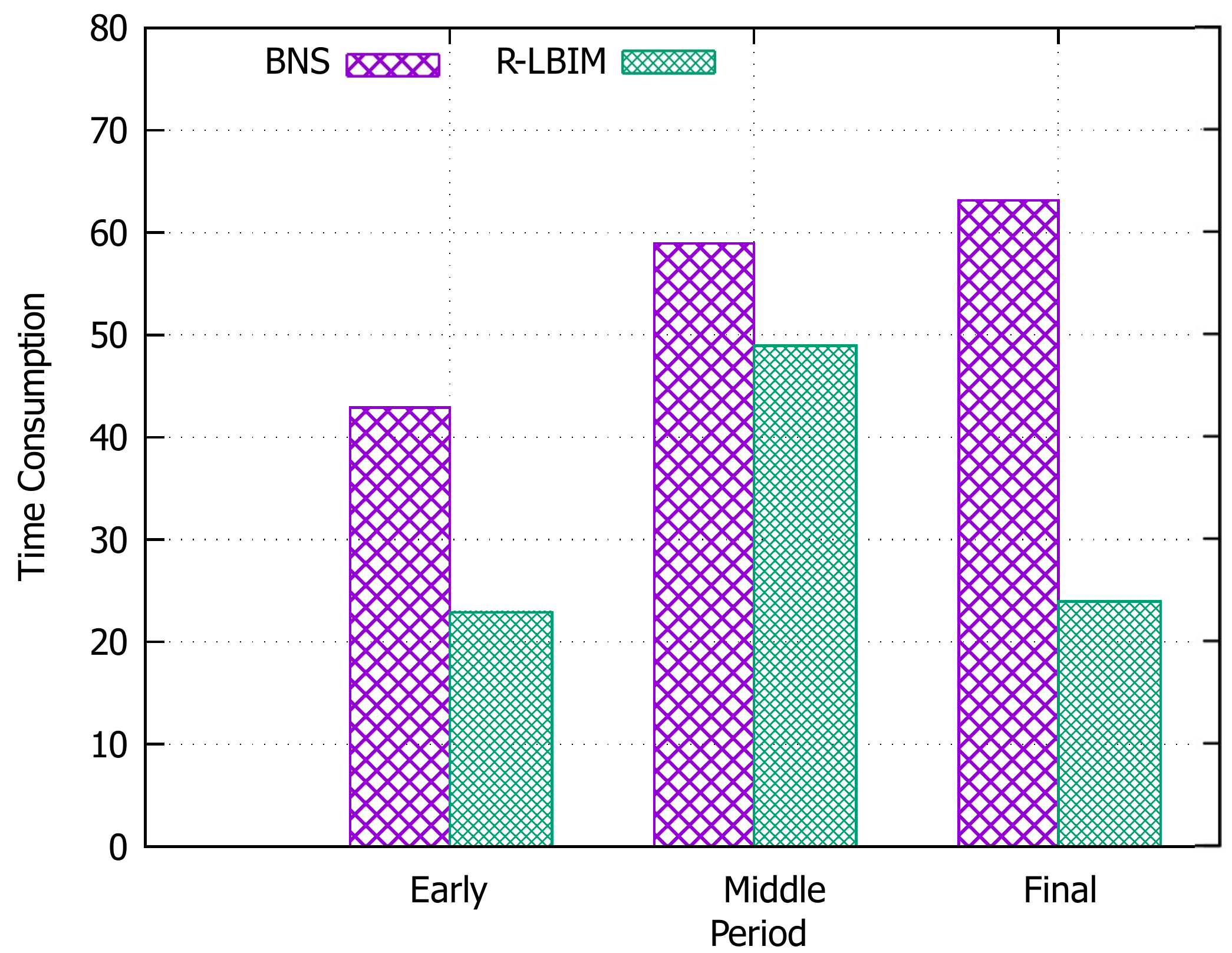}
\\{\small ($b$)~Time Consumption}
\end{minipage}
\caption{BNS v.s. R-LBIM in a single agent setting.}
\label{fig:BNS-SA}
\end{figure*}

\begin{figure*}[ht]
\begin{minipage}[t]{0.5\linewidth}
\centering
\includegraphics[width=2.0in ]{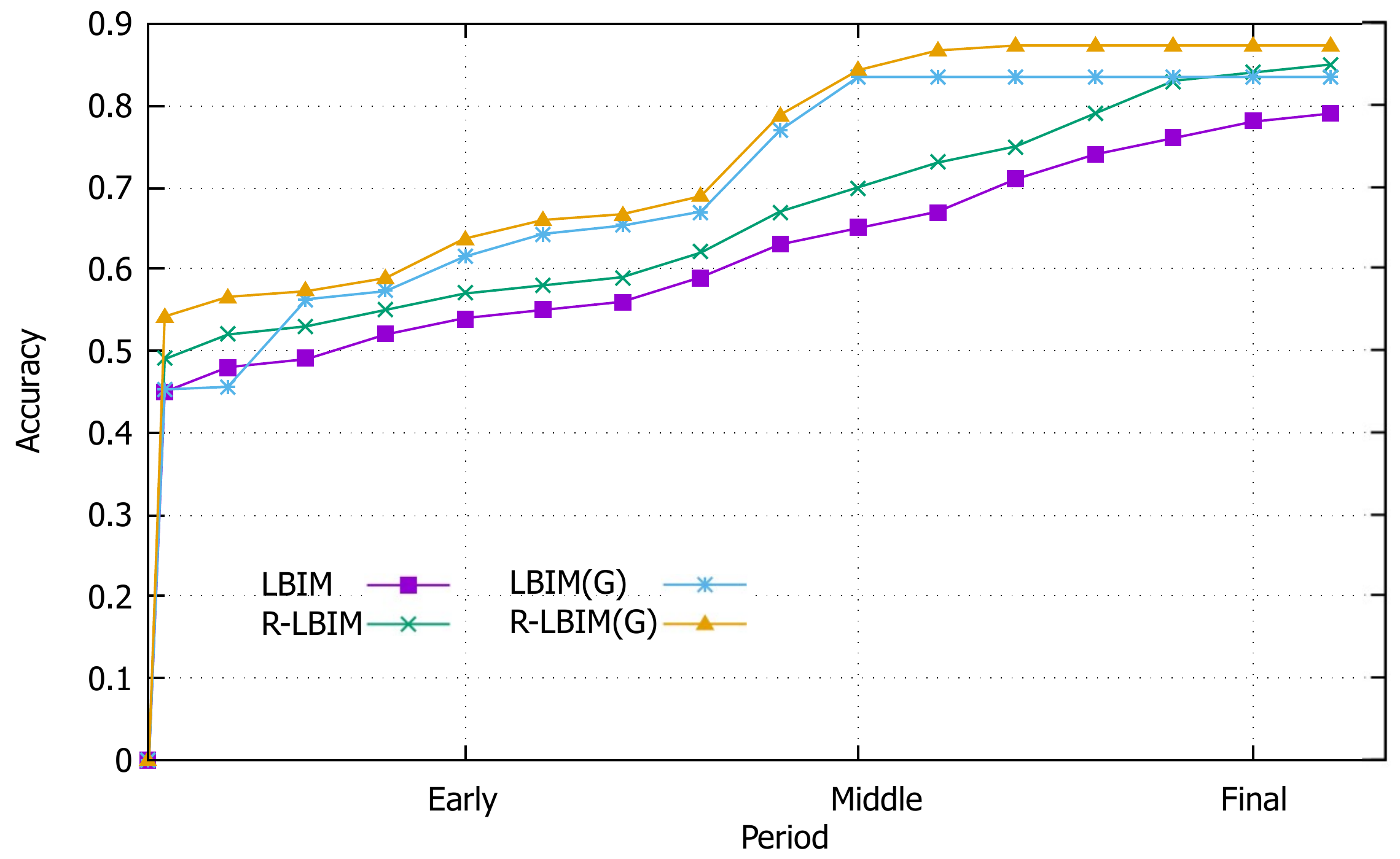}
\\{\small ($a$)~Accuracy Comparison}
\end{minipage}%
\begin{minipage}[t]{0.5\linewidth}
\centering
\includegraphics[width=2.0in ]{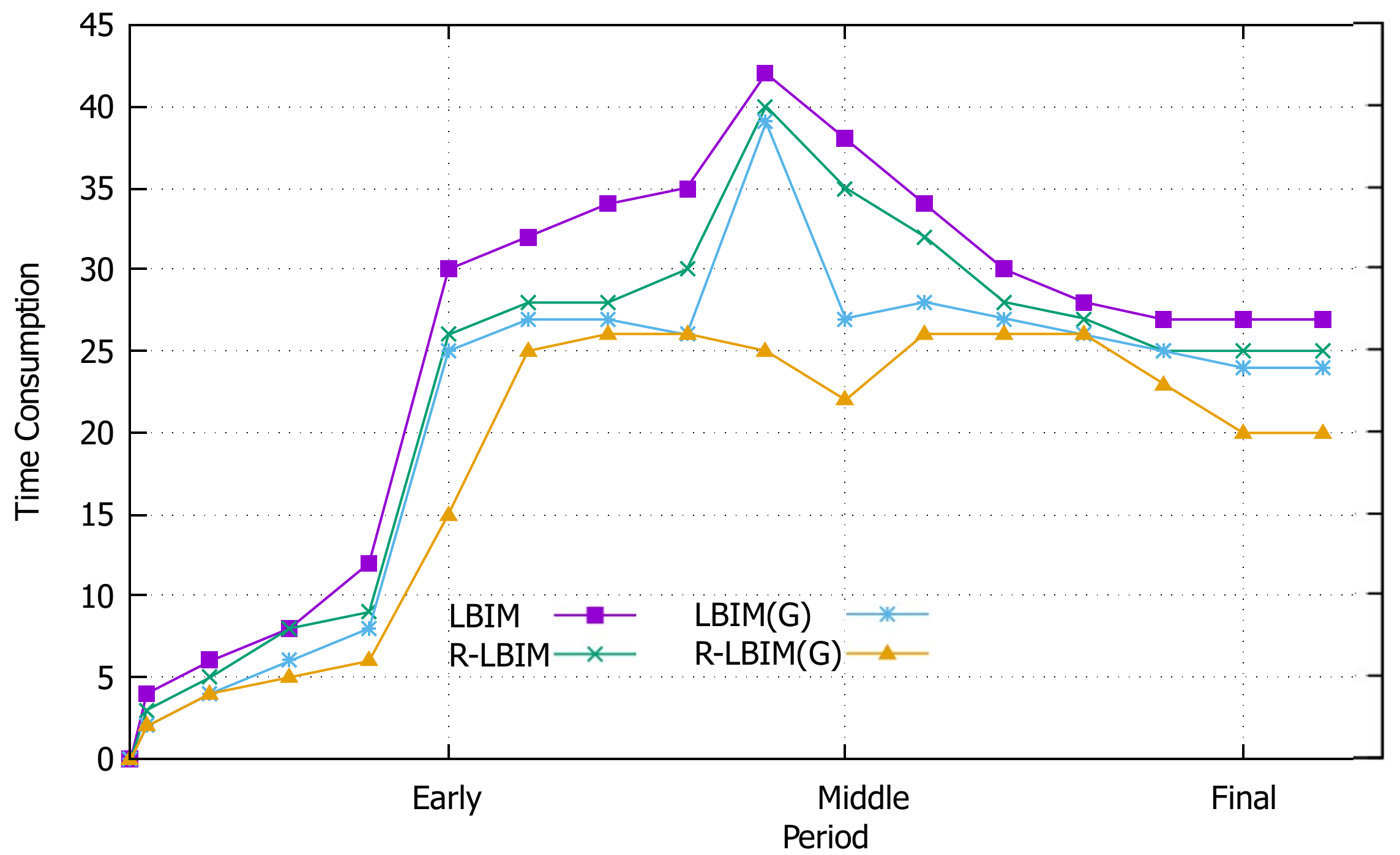}
\\{\small ($b$)~Time Consumption}
\end{minipage}
\caption{Performance of LBIM vs.R-LBIM given a long length of plan sequences in a multi-agent setting.}
\label{fig:LBIM-MA-1}
\end{figure*}

\begin{figure*}[ht]
\begin{minipage}[t]{0.5\linewidth}
\centering
\includegraphics[width=2.0in ]{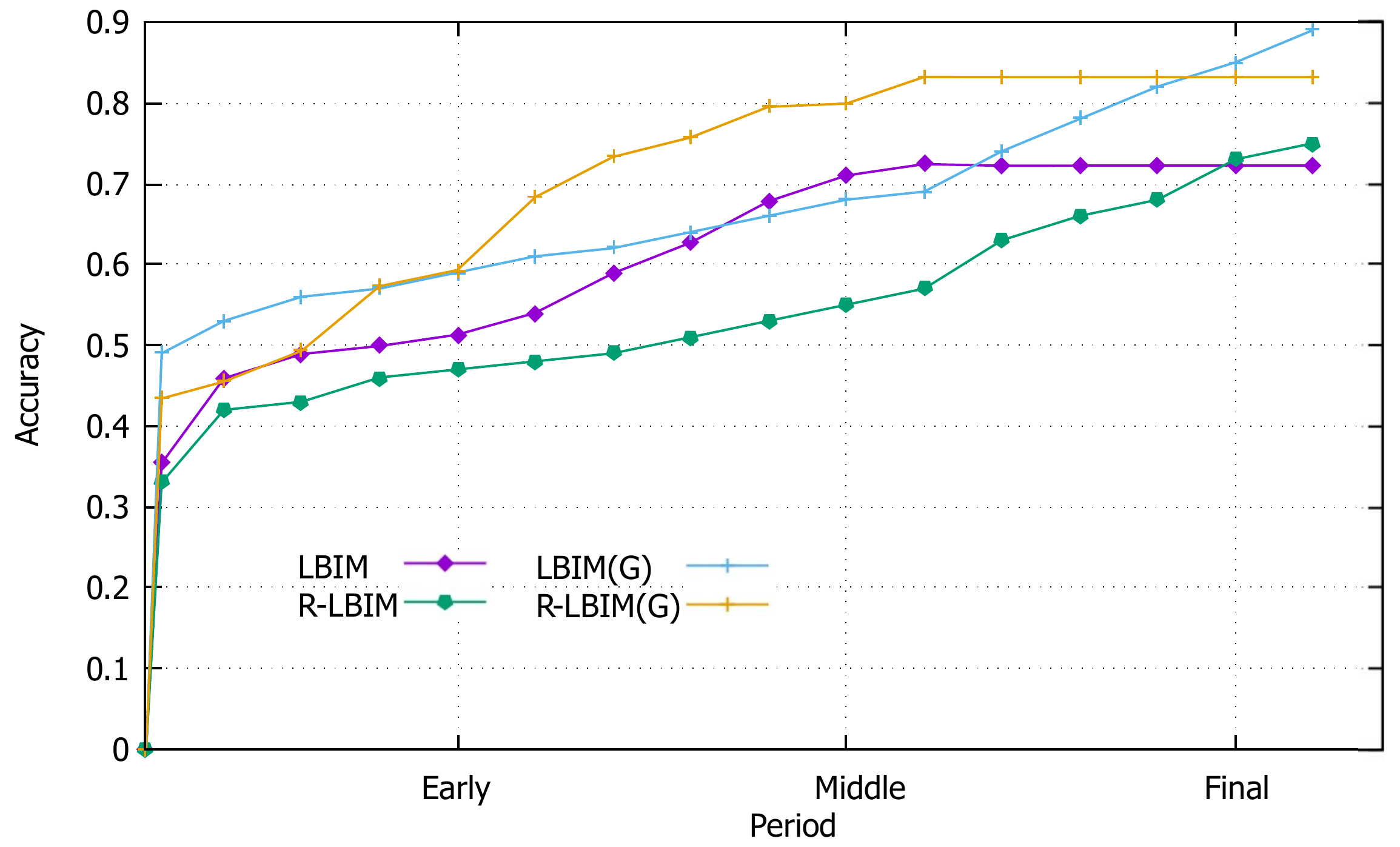}
\\{\small ($a$)~Accuracy Comparison}
\end{minipage}%
\begin{minipage}[t]{0.5\linewidth}
\centering
\includegraphics[width=2.0in ]{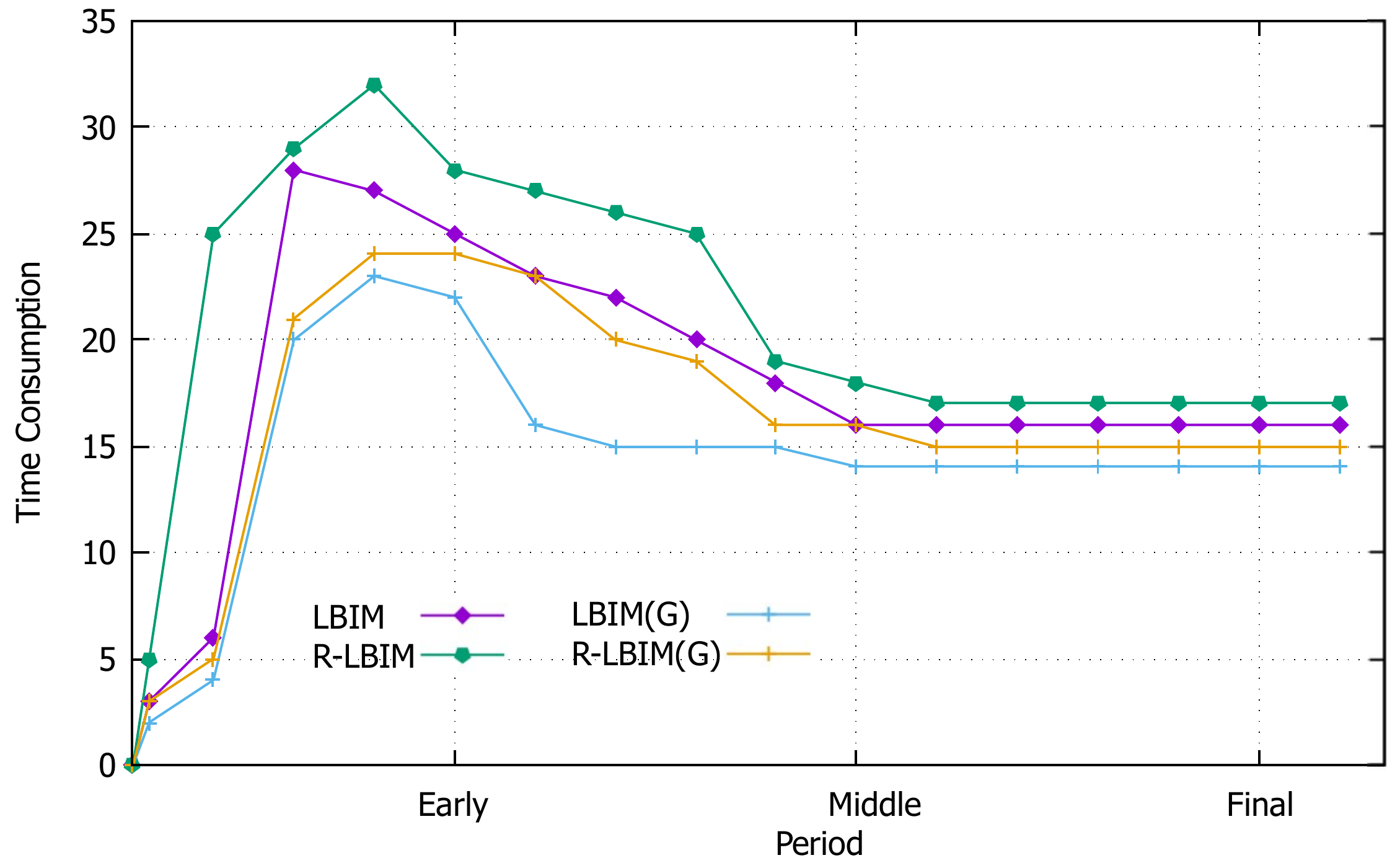}
\\{\small ($b$)~Time Consumption}
\end{minipage}
\caption{Performance of LBIM vs.R-LBIM given a short length of plan sequences in a multi-agent setting.}
\label{fig:LBIM-MA-2}
\end{figure*}

 {\bf [Set 1: Individual Agent Settings]}. \paragraph{We} first compare the performance of BNS and R-LBIM techniques in identifying intentions of individual agents in Fig.~\ref{fig:BNS-SA}. We build the BN models by learning the parameters of conditional probabilistic distributions  from the simulation data. Most of the data are a short length of plan sequences so that the BN models can be built. The BN models fail in the learning and reasoning of the long plan sequence up to a one-digit number of time steps due to the large state space in this problem domain. 
 
In Fig.~\ref{fig:BNS-SA}, we observe that our method R-LBIM outperforms BNS in both accuracy and efficiency. As a type of prescriptive models, BNS demands more computational resources~(times in Fig.~\ref{fig:BNS-SA}($b$)) in a reasoning task of predicting the agents' intentions. On the other hand, R-LBIM has a similar prediction accuracy to BNS. Both methods can perform well in dealing with the short plan sequences although R-LBIM uses a very light model. 
 
{\bf [Set 2: Multi-agent Settings]}. We compare our methods, namely LBIM and R-LBIM, in recognising and grouping intentions of multiple agents. During the intention reconsideration period, every person uses the techniques to identify common intentions for a group of other persons when they are on the way to the first floor. In Fig.~\ref{fig:LBIM-MA-1}, ~\ref{fig:LBIM-MA-2}, we show the performance of LBIM and R-LBIM over both short and long lengths of plan sequences during the simulation. As the simulation involves a large number of persons in the crowd, we first cluster the persons into several groups and then identify the common intentions of the persons who are in the group centre. We denote the LBIM and R-LBIM methods with the grouping preprocess operation as LBIM(G) and R-LBIM(G) respectively.  As expected, both LBIM(G) and R-LBIM(G) spend less time than the methods without the grouping operation at all the cases since they deal with a reduced number of persons in identifying common intentions. In the case of identifying common intentions over short length of plan sequences, LBIM(G) and R-LBIM(G) have similar performance because the behavioural trees that were learned from the data may fully describe the agents' plans and the refinement does not benefit much to the identification. R-LBIM(G) performs better in dealing with the long plan sequences since it uses more informative features to group intentions of multiple agents when their entire planning models are not easy to be built. This observation is also consistent with the performance between LBIM and R-LBIM when dealing with both short and long plan sequences. All of the techniques spend less time in the identification operation in the final stage since most of the persons converge to the intention of arriving the first floor. 

\begin{figure*}[ht]
\begin{minipage}[t]{0.5\linewidth}
\centering
\includegraphics[width=2.0in ]{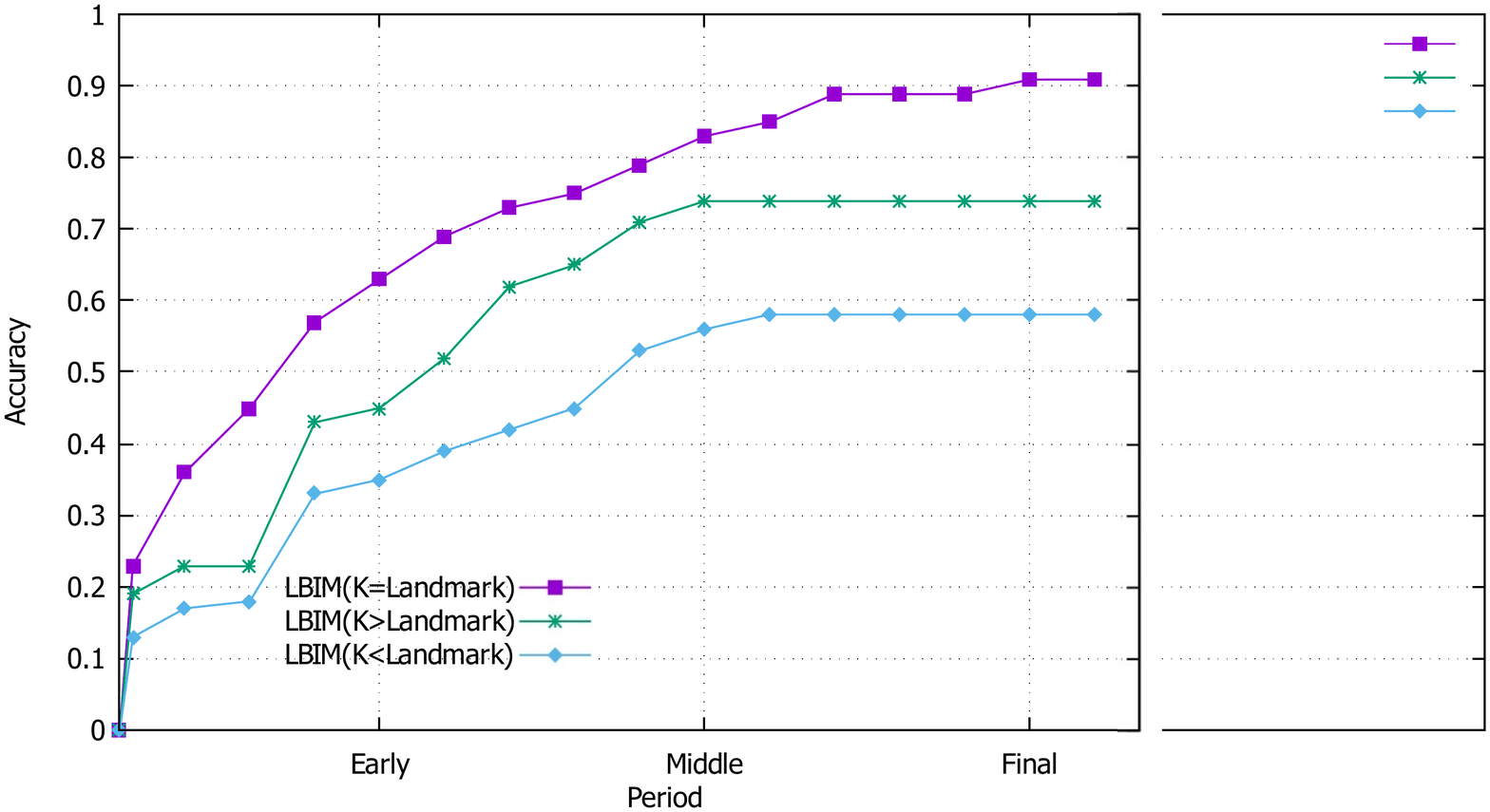}
\\{\small ($a$)~Accuracy Comparison - Long Plan Seqs}
\end{minipage}%
\begin{minipage}[t]{0.5\linewidth}
\centering
\includegraphics[width=2.0in ]{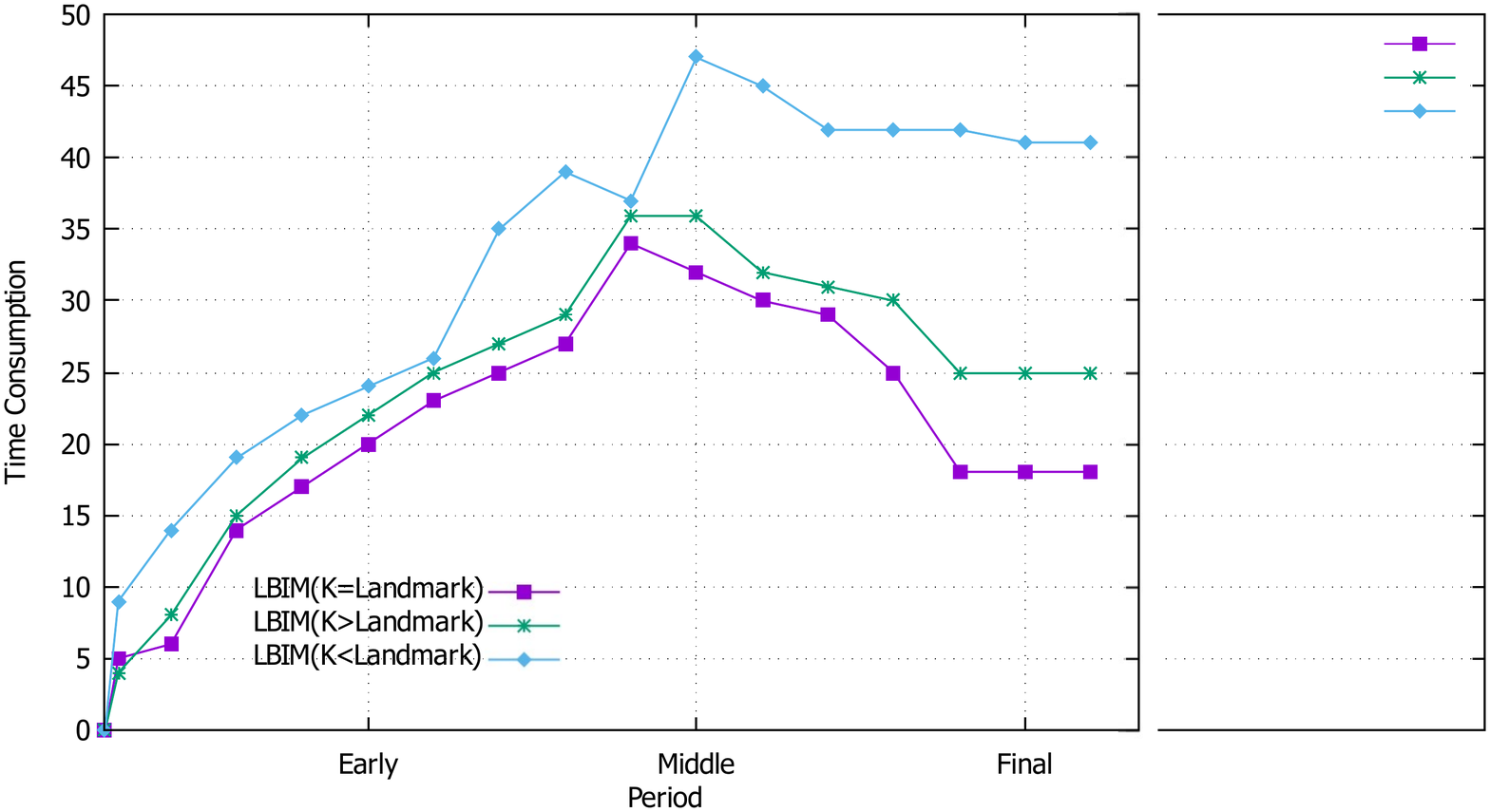}
\\{\small ($b$)~Time Consumption - Long Plan Seqs}
\end{minipage}
\end{figure*}

\begin{figure*}[ht]
\begin{minipage}[t]{0.5\linewidth}
\centering
\includegraphics[width=2.0in ]{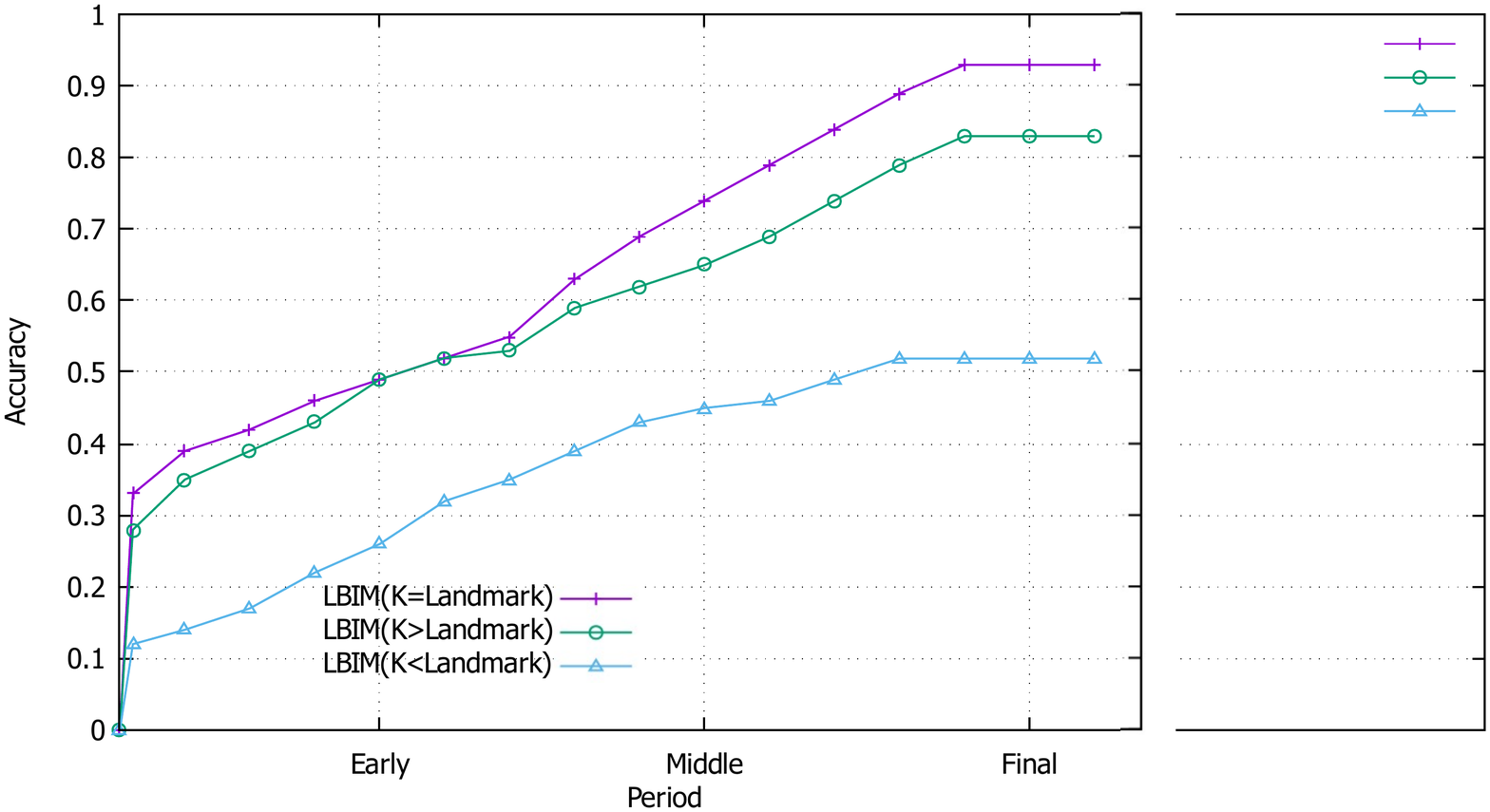}
\\{\small ($a$)~Accuracy Comparison - Short Plan Seqs}
\end{minipage}
\begin{minipage}[t]{0.5\linewidth}
\centering
\includegraphics[width=2.0in ]{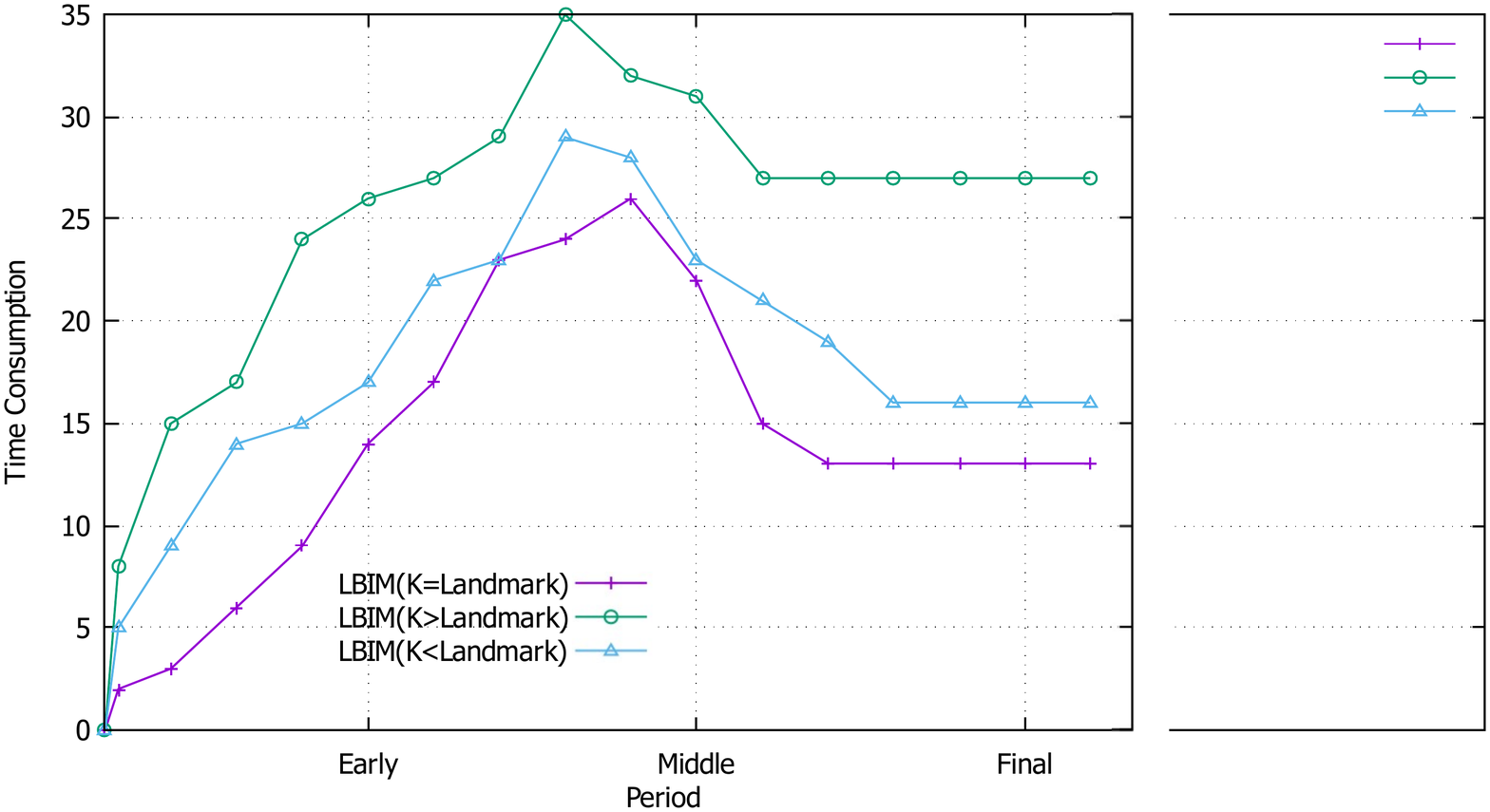}
\\{\small ($b$)~Time Consumption - Short Plan Seqs}
\end{minipage}
\caption{Performance upon different $K$ values in LBIM}
\label{fig:KResults1}
\end{figure*}

\begin{figure*}[ht]
\begin{minipage}[t]{0.5\linewidth}
\centering
\includegraphics[width=2.0in ]{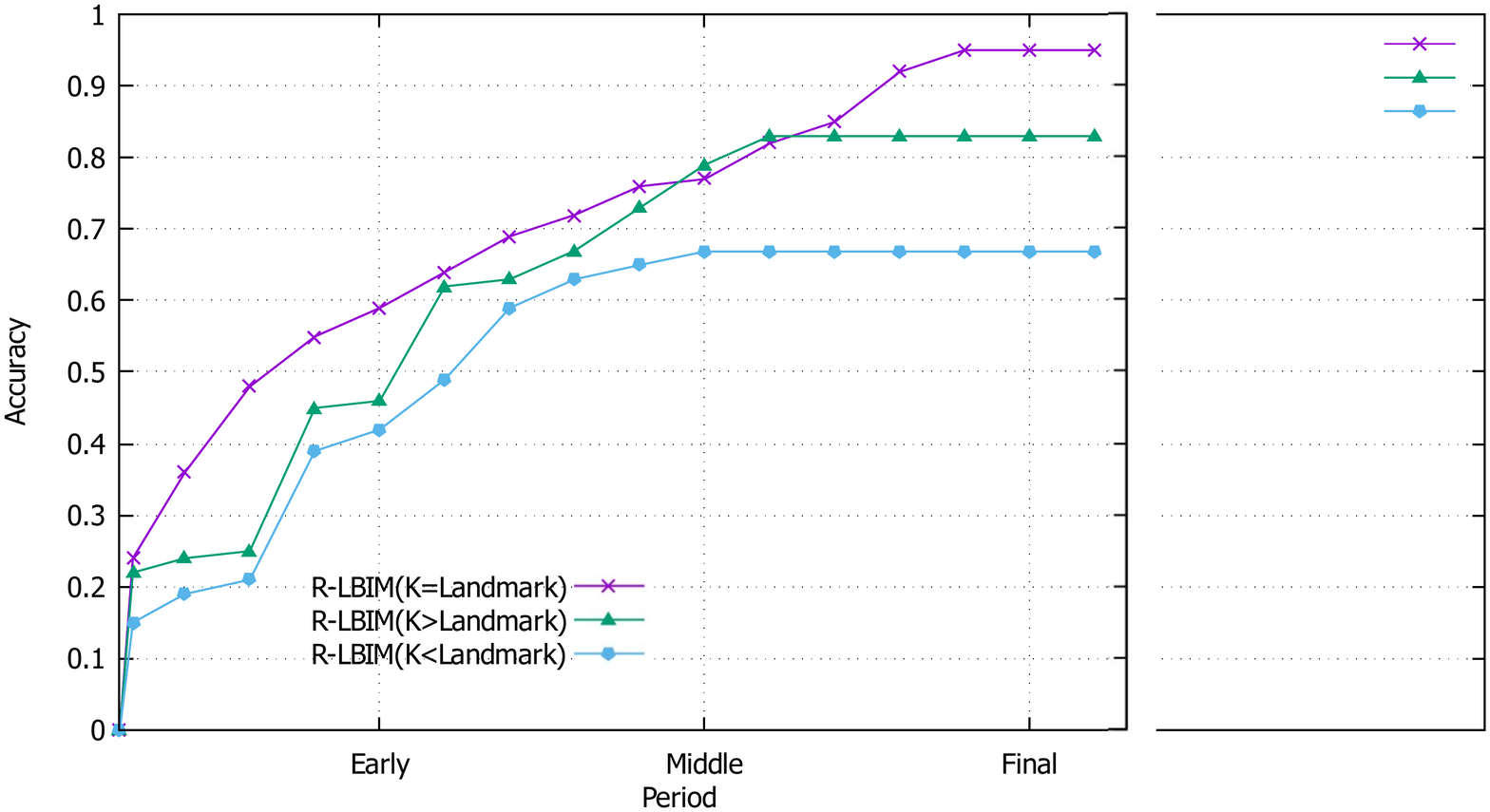}
\\{\small ($a$)~Accuracy Comparison - Long Plan Seqs}
\end{minipage}
\begin{minipage}[t]{0.5\linewidth}
\centering
\includegraphics[width=2.0in ]{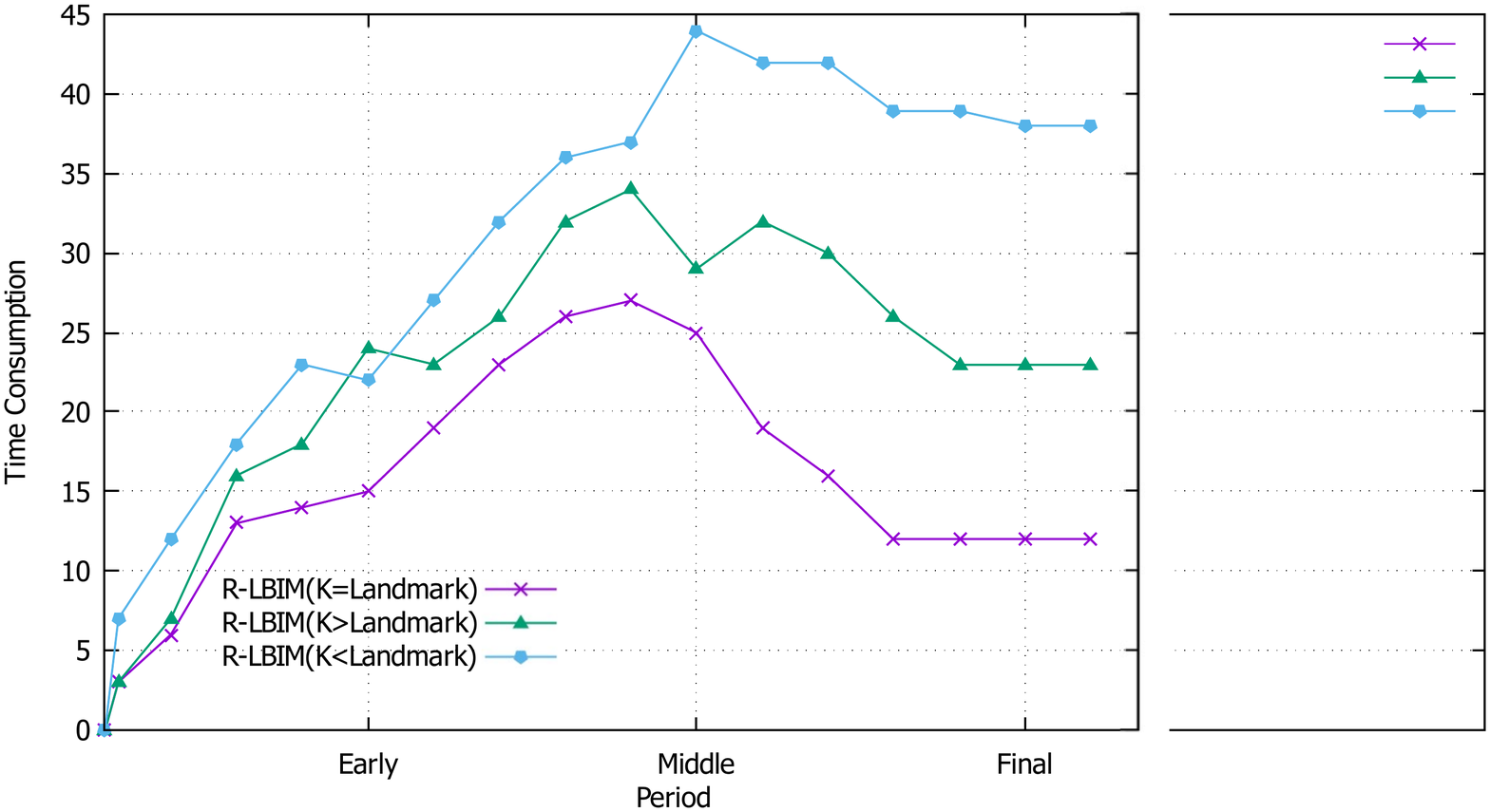}
\\{\small ($b$)~Time Consumption - Long Plan Seqs}
\end{minipage}
\end{figure*}

\begin{figure*}[ht]
\begin{minipage}[t]{0.5\linewidth}
\centering
\includegraphics[width=2.0in ]{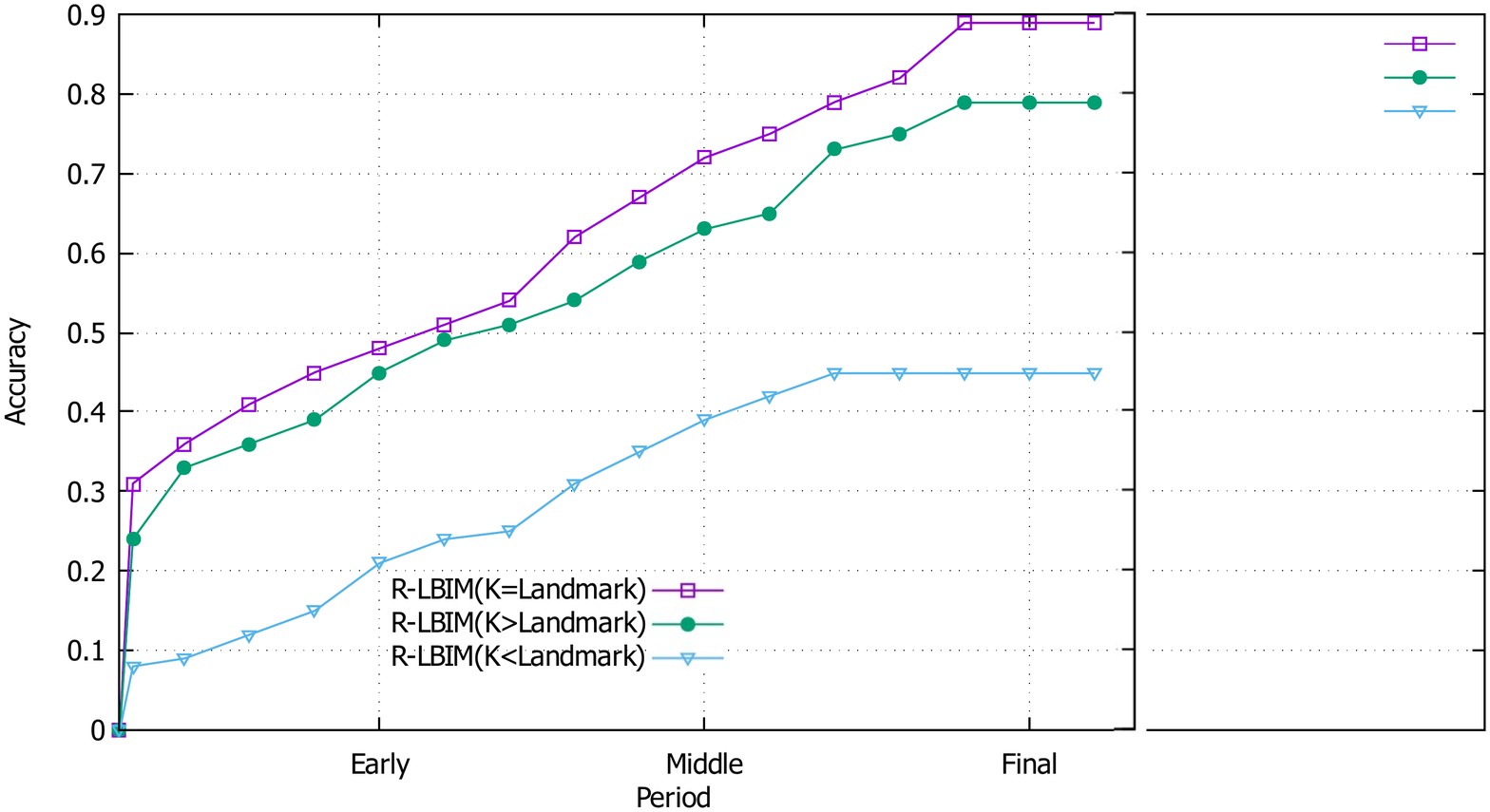}
\\{\small ($a$)~Accuracy Comparison - Short Plan Seqs}
\end{minipage}
\begin{minipage}[t]{0.5\linewidth}
\centering
\includegraphics[width=2.0in ]{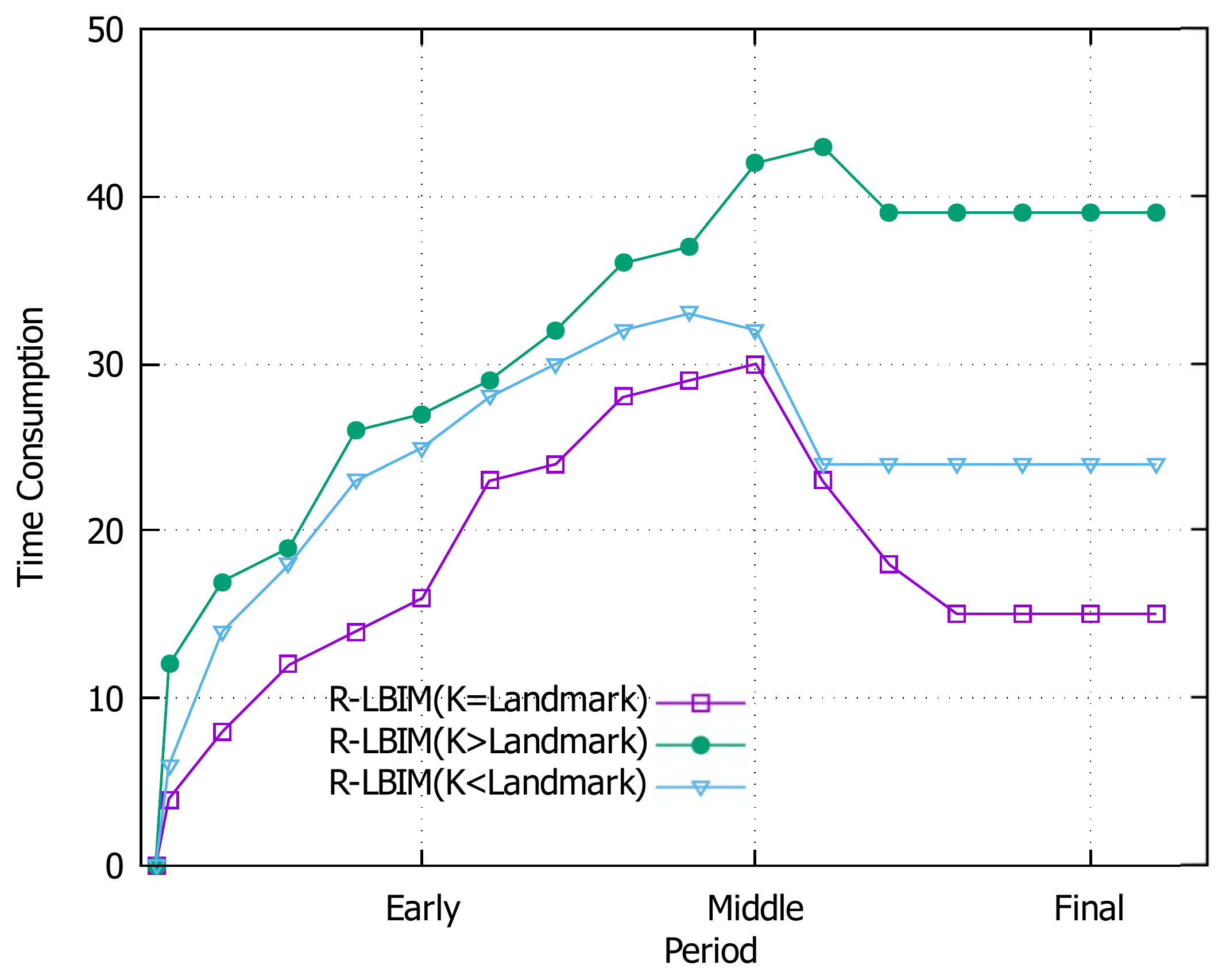}
\\{\small ($b$)~Time Consumption - Short Plan Seqs}
\end{minipage}
\caption{Performance upon different $K$ values in RLBIM - Short Plan Seqs}
\label{fig:KResults2}
\end{figure*}

Finally we conduct another set of experiments to show the performance of the grouping methods when the number of groups, namely the $K$ value, is varied in Algorithm 2. In the previous experiments, we set the $K$ values as the number of most probable landmarks, which is identified automatically in the grouping process. Fig.~\ref{fig:KResults1} and~\ref{fig:KResults2} compare the performance of different $K$ values that are either equal to, larger or smaller than the probably true $K$ values. The methods using the correct $K$ values have much better identification accuracy and consume less times. The improper values lead to more iterations of clustering intentions therefore leading to more computational times. A small $K$ value leads to a low accuracy in identifying common intentions since a person is included in a rather large group of many persons and becomes a random follower in the simulation.  We can observe the consistent results when the $K$ value varies in the LBIM and R-LBIM models.  Hence, it is important to have a good estimation of $K$ values. In our cases, the number of landmarks is a key to decide $K$ values, which fortunately can be gained from domain knowledge as always done in AI planning. 

\section{Related Works}
\label{sec:review}
Intention recognition is closely related to agent decision making and planing in the artificial intelligence research~\cite{georgeff1998belief}\cite{HAJER2020888}. Its research development is interleaved with the research on intelligent agents, decision making, cognitive science and so on.  The initial research has followed a descriptive model based technique particularly with logic-based AI planning techniques~\cite{baral2009probabilistic}. Similar to a traditional AI planner, it lacks the capability of dealing with the uncertainty in an environment. The research further introduced context knowledge into logic programming to improve recognition techniques~\cite{sadri2011logic, sadri2012intention}.  

The majority of intention recognition research has been developed through the applications of Bayesian network~(BN)~\cite{pereira2013state, pereira2013intention}, and focuses on how to conquer the complexity of constructing the models.  Lu{\i}s {\em et al.}~\cite{pereira2012intention} proposed a  three-layer BN model and built the model in an incremental way when more actions are observed over time, which is also classified as one anytime recognition technique~\cite{pereira2010anytime, pereira2009intention}. The approach also involves the integration of context knowledge in the model development~\cite{han2011context}.

Clint~\cite{heinze2004modelling} applied BN fragments to build a recognition model that is easy to be specified and maintained. It has a simple structure and the corresponding knowledge base can be obtained effectively~(automatically designed by domain experts or planned corpus). Moreover, this method may solve the problems of intention change and abandonment. Interestingly, this line of research has focused on applications in a healthy ageing domain. For example, it monitors daily behaviours to detect potential illness of the elderly~\cite{pereira2009elder} and reasons with current intentions of the elderly~\cite{tahboub2006intelligent, heinze2004modelling}. The BN-based intention recognition approach explicitly models uncertain factors impacting intentions and has a great potential in interpret reasons when intentions are not detected or failed in a stochastic setting. However, it demands a vast amount of domain knowledge in order to build a precise model thereby significantly compromising intention recognition performance as reported in our experiments. 

One related work is in the track of plan recognition as intention is considered as one part of plan in an agent model design and development. Geib~\cite{goldman1999new} proposed a hirerarchical and procedural plan recognition model based on probabilistic theory~\cite{charniak1993bayesian}. Through a logical relation of context~\cite{pynadath1995accounting} and a cumulative observation of timing plan, a probability and logical relation of task event assignment in a hierarchical plan~\cite{kautz1986generalized} are calculated for multiple agents. More recently, Mirsky {\em et al.}~\cite{mirsky2019goal, mirsky2018sequential} facilitated plan recognition through  the development of plan equivalence and improved the accuracy by removing the term ambiguity in the plan language. 

The plan recognition model tends to be idealized and agents lack awareness of changes in the surrounding environment and recognition of relations with other agents. Geib~\cite{geib2004assessing} further explored the issue of plan recognition complex planning and behaviour~\cite{armentano2007plan}\cite{8080423}, and built a tree structure to identify a probability distribution of a multi-layer plan~\cite{geib2009probabilistic, geib2003recognizing, banerjee2010multi}. This line of research is still lack of practicability and accuracy of multi-layer plan or multiple plans for a goal. We propose a new model to represent landmark based plan for the multi-agent intention recognition purpose and will complement the existing research in the intention or plan recognition.  

\section{Conclusion}
We present a simple, operational approach for identifying and grouping common intentions of multiple agents. The new LBIM model and its refinement exploits landmark concepts in representing agents' behaviour in a more informative way. The challenge lies in the uncertainty that  agents' full plan can not be obtained before the agents act in an environment. The intention identification becomes difficult as only partial plans are observed  when the agents execute their plan in a real-time mode. We calculate probability distributions of agents over a set of landmarks as an indication of agents'  tendency towards intentions and group their intentions accordingly.  We empirically study the new techniques in a couple of problem domains and show expected performance in different settings. 

This work is the first attempt at dealing with intention recognition for multiple agents. The new approach lacks a sophisticated design in developing relations between plans and intentions, which leads to difficulty in interpreting the intention failure in the setting. In addition, our approach considers a fully observable environment where states and actions can be seen by agents. The model demands more development in a partially observable environment particularly in the learning of behaviour tree~\cite{conroy2015learning}. We will investigate an incremental approach to build the model while identifying agents'  intentions. The interleaved development expects to plan the agents better in a practical setting. 





\bibliographystyle{elsarticle-num}
\bibliography{LBIM}

\begin{thebibliography}{10}
\expandafter\ifx\csname url\endcsname\relax
  \def\url#1{\texttt{#1}}\fi
\expandafter\ifx\csname urlprefix\endcsname\relax\def\urlprefix{URL }\fi
\expandafter\ifx\csname href\endcsname\relax
  \def\href#1#2{#2} \def\path#1{#1}\fi

\bibitem{wooldridge2009introduction}
M.~Wooldridge, {An introduction to multiagent systems}, 2009.

\bibitem{DBLP:journals}
A.~T. Lin, M.~J. Debord, K.~Estabridis, G.~A. Hewer, S.~J. Osher, {CESMA:}
  centralized expert supervises multi-agents, CoRR abs/1902.02311.

\bibitem{pereira2013context}
L.~M. Pereira, {others}, {Context-Dependent Incremental Decision Making
  Scrutinizing the Intentions of Others via Bayesian Network Model
  Construction}, Intelligent Decision Technologies (IDT) 7~(4) (2013) 293--317.

\bibitem{han2011context}
T.~A. Han, L.~M. Pereira, {Context-Dependent Incremental Intention Recognition
  Through Bayesian Network Model Construction}, in: Proceedings of the 8th UAI
  Bayesian Modeling Applications Workshop (UAI-AW), Barcelona,ES, 2011, pp.
  50--58.

\bibitem{stone2020broader}
P.~Stone, {A Broader, More Inclusive Definition of AI}, Journal of Artificial
  General Intelligence (JAGI) 11~(2) (2020) 63--65.

\bibitem{perrault2019ai}
A.~Perrault, F.~Fang, A.~Sinha, M.~Tambe, {AI for Social Impact: Learning and
  Planning in the Data-to-Deployment Pipeline}, ArXiv Preprint ArXiv:2001.00088
  (arXiv).

\bibitem{bellman1957markovian}
R.~Bellman, {A Markovian Decision Process}, Journal of Mathematics and
  Mechanics (JMM) (1957) 679--684.

\bibitem{heinze2004modelling}
C.~Heinze, Modelling intention recognition for intelligent agent systems, Tech.
  rep., Defence Science and Technology Organisation Salisbury (Australia)
  Systems (2004).

\bibitem{pereira2010anytime}
L.~M. Pereira, et~al., Anytime intention recognition via incremental bayesian
  network reconstruction, in: 2010 AAAI Fall Symposium (AAAI), Virginia,USA,
  2010, pp. 20--25.

\bibitem{georgeff1998belief}
M.~Georgeff, B.~Pell, M.~Pollack, M.~Tambe, M.~Wooldridge, {The
  Belief-Desire-Intention Model of Agency}, in: International Workshop on Agent
  Theories, Architectures, and Languages (ATAL), Rhode Island,USA, 1998, pp.
  1--10.

\bibitem{doi:10.2514}
J.~Votion, T.~Feng, Y.~Cao, Risk-Aware Multi-Agent Path Planning for Target
  Detection: A Multi-Agent Reinforcement Learning Approach.

\bibitem{10.1007}
P.~L{\'o}pez~Diez, J.~V. Sundgaard, Facial and cochlear nerves characterization
  using deep reinforcement learning for landmark detection, in: Medical Image
  Computing and Computer Assisted Intervention (MICCAI) 2021, 2021, pp.
  519--528.

\bibitem{hoffmann2004ordered}
J.~Hoffmann, J.~Porteous, L.~Sebastia, {Ordered Landmarks in Planning}, Journal
  of Artificial Intelligence Research (JAIR) 22 (2004) 215--278.

\bibitem{conroy2015learning}
R.~Conroy, Y.~Zeng, M.~Cavazza, Y.~Chen, {Learning Behaviors in Agents Systems
  with Interactive Dynamic Influence Diagrams}, in: Proceedings of the 24th
  International Conference on Artificial Intelligence  (IJCAI), Buenos Aires,
  AR, 2015, pp. 39 -- 45.

\bibitem{ogston2003method}
E.~Ogston, B.~Overeinder, M.~Van~Steen, F.~Brazier, {A Method for Decentralized
  Clustering in Large Multi-Agent Systems}, in: Proceedings of the 2nd
  International Joint Conference on Autonomous Agents and Multiagent Systems
  (IJCAI), Melbourne,AU, 2003, pp. 789--796.

\bibitem{sinaga2020unsupervised}
K.~P. Sinaga, M.-S. Yang, Unsupervised k-means clustering algorithm, IEEE
  Access 8 (2020) 80716--80727.

\bibitem{mesbahi2015multi}
N.~Mesbahi, O.~Kazar, S.~Benharzallah, M.~Zoubeidi, S.~Bourekkache,
  Multi-agents approach for data mining based k-means for improving the
  decision process in the erp systems, International Journal of Decision
  Support System Technology (IJDSST) 7~(2) (2015) 1--14.

\bibitem{coma2003multi}
R.~Coma, G.~Simon, M.~Coletta, {A Multi-Agent Architecture for Agents
  Clustering}, in: Proceedings of 4th International Workshop on Agent-Based
  Simulation (ABS), Boston,USA, 2003.

\bibitem{lees2002history}
M.~Lees, A history of the tileworld agent testbed, Computer Science Technical
  Report No. NOTTCSWP-2002 (CSTR) 1 (2002) 2001--2002.

\bibitem{zeng2017using}
Y.~Zeng, Z.~Zhang, T.~A. Han, I.~R. Spears, S.~Qin, {Using Intention
  Recognition in A Simulation Platform to Assess Physical Activity Levels of An
  Office Building}, in: Proceedings of the 16th Conference on Autonomous Agents
  and Multi-Agent Systems (AAMAS), Sao Paulo,BR, 2017, pp. 1817--1819.

\bibitem{HAJER2020888}
B.~Hajer, B.~Arwa, H.~Lobna, G.~Khaled, Intention mining data preprocessing
  based on multi-agents system, Procedia Computer Science 176 (2020) 888--897.

\bibitem{baral2009probabilistic}
C.~Baral, M.~Gelfond, N.~Rushton, {Probabilistic Reasoning with Answer Sets},
  Theory and Practice of Logic Programming (TPLP) 9~(1) (2009) 57--144.

\bibitem{sadri2011logic}
F.~Sadri, {Logic-Based Approaches to Intention Recognition}, in: Handbook of
  Research on Ambient Intelligence and Smart Environments: Trends and
  Perspectives, 2011, pp. 346--375.

\bibitem{sadri2012intention}
F.~Sadri, {Intention Recognition in Agents for Ambient Intelligence:
  Logic-Based Approaches}, Agents and Ambient Intelligence (AAI) 12 (2012)
  197--236.

\bibitem{pereira2013state}
Pereira, L.~Moniz, {State-of-The-Art of Intention Recognition and its Use in
  Decision Making}, AI Communications (AIC) 26~(2) (2013) 237--246.

\bibitem{pereira2013intention}
Pereira, L.~Moniz, Intention-based decision making via intention recognition
  and its applications, Human Behavior Recognition Technologies: Intelligent
  Applications for Monitoring and Security (IGI Global) (2013) 174--211.

\bibitem{pereira2012intention}
L.~M. Pereira, Intention Recognition, Commitment and Their Roles in the
  Evolution of Cooperation, 2012.

\bibitem{pereira2009intention}
L.~M. Pereira, {others}, {Intention Recognition via Causal Bayes Networks Plus
  Plan Generation}, in: Proceedings of the 14th Portuguese Conference on
  Artificial Intelligence: Progress in Artificial Intelligence (PCAI),
  Portugal,PT, 2009, pp. 138--149.

\bibitem{pereira2009elder}
L.~M. Pereira, {others}, {Elder Care via Intention Recognition and Evolution
  Prospection}, in: Proceedings of the 18th International Conference on
  Applications of Declarative Programming and Knowledge Management(ADPKM),
  Portugal,PT, 2009, pp. 170--187.

\bibitem{tahboub2006intelligent}
K.~A. Tahboub, {Intelligent Human-Machine Interaction Based on Dynamic Bayesian
  Networks Probabilistic Intention Recognition}, Journal of Intelligent and
  Robotic Systems (JIRS) 45~(1) (2006) 31--52.

\bibitem{goldman1999new}
R.~P. Goldman, C.~W. Geib, C.~A. Miller, {A New Model of Plan Recognition}, in:
  Proceedings of the 15th Conference on Uncertainty in Artificial Intelligence
  (UAI), Stockholm,SV, 1999, pp. 245--254.

\bibitem{charniak1993bayesian}
E.~Charniak, R.~P. Goldman, {A Bayesian Model of Plan Recognition}, Artificial
  Intelligence (AI) 64~(1) (1993) 53--79.

\bibitem{pynadath1995accounting}
D.~V. Pynadath, M.~P. Wellman, {Accounting for Context in Plan Recognition,
  with Application to Traffic Monitoring}, in: Proceedings of the 11th
  Conference on Uncertainty in Artificial Intelligence (CUAI), Paris,France,
  1995, pp. 472--481.

\bibitem{kautz1986generalized}
H.~A. Kautz, J.~F. Allen, Generalized plan recognition, in: Proceedings of the
  5th AAAI National Conference on Artificial Intelligence (NCAI),
  Philadelphia,USA, 1986, pp. 32--37.

\bibitem{mirsky2019goal}
R.~Mirsky, K.~Gal, R.~Stern, M.~Kalech, {Goal and Plan Recognition Design for
  Plan Libraries}, ACM Transactions on Intelligent Systems and Technology
  (TIST) 10~(2) (2019) 1--23.

\bibitem{mirsky2018sequential}
R.~Mirsky, R.~Stern, K.~Gal, M.~Kalech, {Sequential Plan Recognition: An
  Iterative Approach to Disambiguating Between Hypotheses}, Artificial
  Intelligence (AI) 260 (2018) 51--73.

\bibitem{geib2004assessing}
C.~W. Geib, Assessing the complexity of plan recognition, in: Proceedings of
  the 19th National Conference on Artifical Intelligence (NCAI),
  California,USA, 2004, pp. 507--512.

\bibitem{armentano2007plan}
M.~G. Armentano, A.~Amandi, {Plan Recognition for Interface Agents}, Artificial
  Intelligence Review (AIR) 28~(2) (2007) 131--162.

\bibitem{8080423}
G.~Grossi, B.~Ross, Evolved communication strategies and emergent behaviour of
  multi-agents in pursuit domains, in: 2017 IEEE Conference on Computational
  Intelligence and Games (CIG), 2017, pp. 110--117.

\bibitem{geib2009probabilistic}
C.~W. Geib, R.~P. Goldman, {A Probabilistic Plan Recognition Algorithm Based on
  Plan Tree Grammars}, Artificial Intelligence (AI) 173~(11) (2009) 1101--1132.

\bibitem{geib2003recognizing}
C.~W. Geib, R.~P. Goldman, {Recognizing Plan/Goal Abandonment}, in: Proceedings
  of the 18th International Joint Conference on Artificial Intelligence
  (IJCAI), Acapulco, MX, 2003, pp. 1515--1517.

\bibitem{banerjee2010multi}
B.~Banerjee, L.~Kraemer, J.~Lyle, Multi-agent plan recognition: Formalization
  and algorithms, in: Proceedings of the 24th AAAI Conference on Artificial
  Intelligence (AAAI), Georgia,USA, 2010, pp. 1059--1064.

\end{thebibliography}







\end{document}